\documentclass[journal]{IEEEtran}
\usepackage{amsmath,amsfonts}
\usepackage{algorithmic}
\usepackage{algorithm}
\usepackage{array}
\usepackage[caption=false,font=normalsize,labelfont=sf,textfont=sf]{subfig}
\usepackage{textcomp}
\usepackage{stfloats}
\usepackage{url}
\usepackage{verbatim}
\usepackage{graphicx}
\usepackage{makecell}
\usepackage[comma,sort&compress,numbers]{natbib}
\usepackage[font=small]{caption}
\usepackage{multirow}
\usepackage{booktabs}
\usepackage[table]{xcolor} 
\usepackage{tikz}
\usepackage{hyperref}
\usepackage{subcaption}
\usepackage{etoolbox}
\usepackage{caption}  
\usepackage{soul}
\usetikzlibrary{positioning,calc,fit}

\newcommand{\name}{ViVo}

\newcolumntype{P}[1]{>{\centering\arraybackslash}p{#1}}
\newcommand{\rightdownarrow}{\tikz[baseline=-1ex]{\draw[->] (0,0) -- (0,-0.1) -- (0.3,-0.1);}}


\begin{document}

\title{\name: A Dataset for Volumetric Video Reconstruction and Compression}

\author{Adrian Azzarelli,~\IEEEmembership{Student Member,~IEEE,}
Ge Gao,~\IEEEmembership{Member,~IEEE,}
Ho Man Kwan,~\IEEEmembership{Student Member,~IEEE,}\\
Fan Zhang,~\IEEEmembership{Senior Member,~IEEE,}
Nantheera Anantrasirichai,~\IEEEmembership{Member,~IEEE,}\\
Ollie Moolan-Feroze,
David Bull,~\IEEEmembership{Fellow,~IEEE}

\thanks{Manuscript submitted in May 2025.}
\thanks{Adrian Azzarelli, Ge Gao, Ho Man Kwan, Fan Zhang, Nantheera Anantrasirichai and David Bull are with the Visual Information Lab, University of Bristol, Bristol, BS1 5DD, U.K. (e-mail: \{a.azzarelli, ge1.gao, hm.kwan, fan.zhang, n.anantrasirichai, dave.bull\}@bristol.ac.uk). Ollie Moolan-Feroze is with Condense Reality Ltd, 1 Canon’s Road, Bristol, BS1 5TX, U.K. (e-mail: ollie@condensereality.com)}

\thanks{The authors acknowledge the funding from Innovate UK (10107058) and the UKRI MyWorld Strength in Places Programme (SIPF00006/1).}
}

\maketitle

\begin{abstract}
As research on neural volumetric video reconstruction and compression flourishes, there is a need for diverse and realistic datasets, which can be used to develop and validate reconstruction and compression models. However, existing volumetric video datasets lack diverse content in terms of both semantic and low-level features that are commonly present in real-world production pipelines. In this context, we propose a new dataset, \name, for VolumetrIc VideO reconstruction and compression. The dataset is faithful to real-world volumetric video production and is the first dataset to extend the definition of diversity to include both human-centric characteristics (skin, hair, etc.) and dynamic visual phenomena (transparent, reflective, liquid, etc.). Each video sequence in this database contains raw data including fourteen multi-view RGB and depth video pairs, synchronized at 30FPS with per-frame calibration and audio data, and their associated 2-D foreground masks and 3-D point clouds. To demonstrate the use of this database, we have benchmarked three state-of-the-art (SotA) 3-D reconstruction methods and two volumetric video compression algorithms. The obtained results evidence the challenging nature of the proposed dataset and the limitations of existing datasets for both volumetric video reconstruction and compression tasks, highlighting the need to develop more effective algorithms for these applications. The database and the associated results are available at \url{https://vivo-bvicr.github.io/}
\end{abstract}


\section{Introduction}\label{sec: introduction}
\IEEEPARstart{A}{acquiring} diverse and representative volumetric video data is an active challenge that is key to computer vision research. Over the years, datasets such as ZJU MoCap and CMU Panoptic \cite{peng2021zju, joo2015panoptic} have facilitated advances in numerous fields, including dynamic 3-D reconstruction and multi-view video (MVV) compression. Consequently, as research related to creative volumetric video reconstruction and compression emerges, the demand for effective datasets also grows.

The volumetric video production pipeline begins with MVV capture, where videos and other sensor data are collected from various viewing angles and positions around a target scene (e.g. a performer). Based on the captured data, a realistic dynamic 3-D scene is then synthesized, compressed, and transmitted to an audience for consumption. However, most existing MVV datasets focus on semantic scenarios rather than on different low-level materials and texture types (e.g. dynamic textures and transparent materials), which are important for assessing performance on the target tasks (reconstruction or compression). Many datasets also do not simulate the real volumetric video production scenarios. As they focus only on human-centric volumetric reconstruction, the capture area is designed to physically isolate the actor from the film crew to minimize background noise. In real production scenarios, physically isolating the actor is not practical as it impacts the quality of the performance. Hence, removing/minimizing dynamic background noise (e.g., the film crew moving around) is a major challenge that is not well represented by existing datasets.

        


\begin{figure*}
    \centering
    \begin{tikzpicture}
        \node[anchor=south west,inner sep=0] (image) at (0,0) {\includegraphics[width=1.\linewidth]{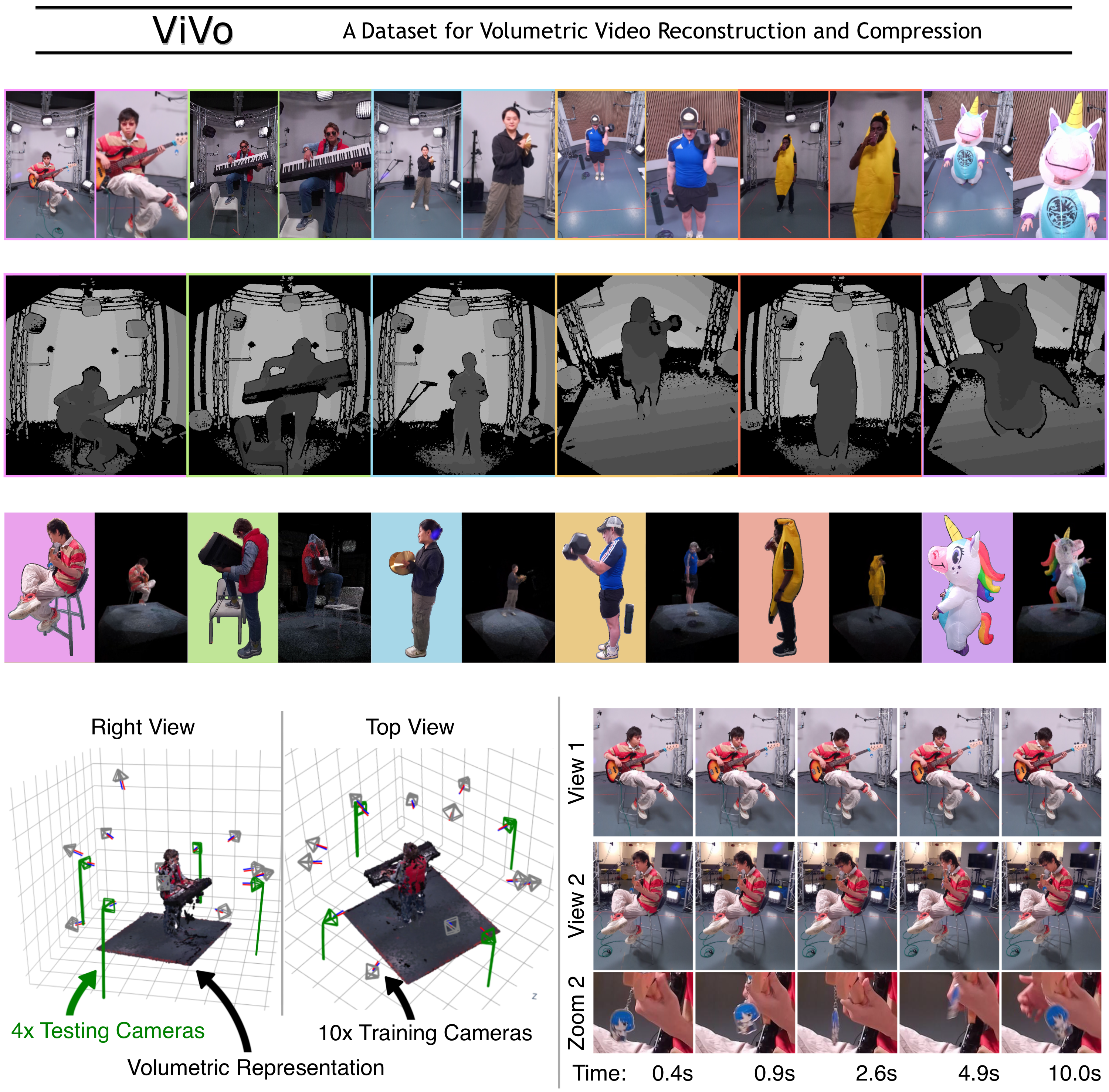}};
        
        \node at ($(image.south west) + (2.55,16.65)$) {\textcolor{black}{(a) RGB Data + Zoomed Image}};
        \node at ($(image.south west) + (1.3,13.65)$) {\textcolor{black}{(b) Depth Data}};
        \node at ($(image.south west) + (2.35,9.75)$) {\textcolor{black}{(c) Mask + Point Cloud Data}};
        \node at ($(image.south west) + (3.3,6.6)$) {\textcolor{black}{(d) Camera Calibration + Train-Test Split}};
        \node at ($(image.south west) + (12.,6.6)$) {\textcolor{black}{(e) Multi-view Sample Sequence}};
    \end{tikzpicture}
    \caption{We showcase the scenes and relevant data associated with the NVS and MVV compression experiments. (a-b) is the source data. (c) is the supplementary data. (d) shows the camera poses and indicates in green the position of the test cameras. (e) illustrates a sample sequence from two cameras}
    \label{fig:mainfig}
\end{figure*}

To further drive research linked to \textbf{V}olumetr\textbf{i}c \textbf{V}ide\textbf{o} reconstruction and compression, this paper contributes \textbf{\name}, a diverse and professionally collected dataset of MVV sequences that are faithful to the volumetric media production process and capture a high variety of dynamic objects, people, textures and practical effects. This database contains 32 unique volumetric video sequences, lasting between 60 and 120 seconds. Per sequence, each frame is associated with 14 RGB views at a spatial resolution of 2560$\times$1440 pixels, and synchronized 14 depth views at 650$\times$576 pixels, as well as their corresponding camera poses, intrinsic data and audio signals (as shown in \autoref{fig:mainfig}). These metadata are important in real production applications but are not supported in existing datasets. The collected content reflects challenges that are present in both the real and research settings, in particular including various dynamic scenes such as fire, liquid, inflatable costumes, and balloons. As the dataset focuses on human activities, \name \ also captures actors with different ages, genders, hair styles, skin colors, and clothes. Based on the collected data, we also provide synthesized per-frame colored point clouds (up to $155$ million points per second), dynamic 2-D foreground masks, and 3-D box masks for various processing tasks. To ensure high-quality 2-D masks are generated, we integrate SAM2 \cite{ravi2024sam} into a GUI that allows users to re-synthesize our masks or create their own, as well as apply erosion and dilation. Manually drawn static masks are also provided and segment the staging area from the dynamic background.

\IEEEpubidadjcol

To demonstrate the use of this dataset for volumetric video reconstruction and compression, we benchmarked three state-of-the-art (SotA) models on novel view synthesis (NVS) and three SotA methods for MVV compression. Contrary to findings in studies that evaluate dense-view NVS, the sparse-view NVS experiments show that newer approaches fail to capture fine details both spatially and temporally. The results also indicate that classical image quality metrics are not effective for evaluating visual performance as they are prone to temporal bias. The experimental findings in MVV compression reveal that the evaluated baseline methods struggle to reconstruct fine-grained or highly dynamic textures accurately - an issue that becomes particularly pronounced under low-bitrate conditions.

The main contributions of this work are summarized as follows.
\begin{itemize}
    \item As far as we are aware, ViVo is the \textbf{first} multi-view human video dataset which simulates real production settings, and covers a diverse range of people and dynamic visual phenomena. It is also the \textbf{only} dataset to provide per-frame calibration, RGB and depth data for each captured scenario.
    \item Associated with this dataset, we further provide a set of \textbf{open-source data processing tools} that allow users to: (i) utilize a SotA video segmentation method to produce high quality per-frame 2-D dynamic masks, and (ii) generate colored point clouds with up to $155$ million points per frame. 
    \item We conduct experiments that benchmark both SotA dynamic NVS models and SotA MVV compression models on the new database. A detailed analysis is provided on the benchmark results which identifies potential future research directions.
\end{itemize}

The remainder of this paper is organized as follows. \autoref{sec:literature} briefly summarizes the related work in volumetric video reconstruction and compression, in particular focusing on existing databases for these tasks. \autoref{sec:dataset} provides a detailed description of the database, including captured scenarios, acquisition configurations, and the associated data and tools. Based on the dataset, the designed benchmark experiments are described in \autoref{sec: experiment design} for both NVS and MVV tasks, and the generated results are then discussed in \autoref{sec: benchmarks}. Finally, \autoref{sec:conclusion} offers a summary of this paper and outlines future work.

\section{Related Works}
\label{sec:literature}

In this section, we first review the previous work on volumetric video reconstruction and commonly used databases. We then provide a summary of the notable works on volumetric video compression and their test datasets. 

\subsection{Volumetric Video Reconstruction} \label{sec: dnvs.models}
Volumetric video reconstruction, also known as dynamic 3-D reconstruction, aims to create a photorealistic representation of a dynamic 3-D scene based on 2-D MVV observations. Major advancements in this task are predominantly driven by neural radiance fields (NeRF) \cite{mildenhall2021nerf,muller2022instant,kwon2021neural} and Gaussian splatting (GS) \cite{kerbl20233d,moreau2024human} research. Methods focusing on volumetric scene reconstruction have overcome prior challenges associated with mesh-based/photogrammetric reconstruction. For dynamic NVS, the accomplishments mainly target learning and disentangling view-dependent visual effects and dynamic motions. Among these, dynamic GS solutions \cite{wu20244d,li2024spacetime, yang2024deformable, huang2024sc} offer significantly faster rendering while also producing better qualitative and quantitative results, compared with NeRF-based alternatives \cite{fridovich2023k, cao2023hexplane}.

A notable approach is 4D-GS~\cite{wu20244d}, which accelerates rendering by modeling temporal deformations of 3D-GS~\cite{kerbl20233d} point parameters using the K-Planes representation~\cite{fridovich2023k} for learning point deformations. 4D-GS offers a simple solution that works well on most reconstruction tasks. However, it remains sensitive to dynamic texture artifacts. To address this, STG~\cite{li2024spacetime} models opacity as a radial basis function, significantly enhancing performance on scenes with dynamic textures. STG also extends the 3D-GS rendering method to directly handle 4-D point properties, achieving much higher frame rates. In contrast, SC-GS~\cite{huang2024sc} achieves SotA results by enforcing volumetric rigidity via a shallow hierarchy of control points fused with linear blended skinning \cite{robert2007lbs} to model control point deformations via warping. Overall, 4D-GS, STG and SC-GS offer unique solutions that represent various approaches to resolving dynamic 3-D reconstruction. We benchmark all three models in our study and discuss the performance in detail in \autoref{sec: benchmarks}.

To facilitate research on volumetric video reconstruction, various datasets have been proposed for algorithm development and benchmarking. It is noted that many existing volumetric videos datasets, such as DNA-Rendering \cite{cheng2023dna} and MVHumanNet \cite{xiong2024mvhumannet}, target dynamic 3-D human reconstruction, which simplifies the reconstruction problem by relying heavily on 2-D masks to avoid background modeling \cite{azzarelli2024exploring}. As a result, these datasets and methods are not as popular or widely used for mainstream 3-D reconstruction as 2-D masks are not trivial to generate or interpret for dynamic textures and semi-transparent materials. Instead, the focus lies on full scene generation. For example, the CMU Panoptic dataset \cite{joo2015panoptic} was used in \cite{yang2024deformable} to evaluate a novel dynamic 3D-GS model. However, the dataset lacks transparent and reflective materials, which is one of the primary goals of dynamic NVS as discussed above. A better example of visual diversity can be seen in the popular DyNeRF dataset \cite{li2022neural}, where dynamic  fire, smoke, and liquid textures are captured in several cooking scenes. However, because the cameras are positioned in a forward-facing configuration,  reconstruction is 2.5-D rather than 3-D. Furthermore, the test camera is also selected from the densest position w.r.t the position of training cameras in the scene. 


\subsection{Volumetric Video Compression}\label{sec: mvv-compression}
Volumetric compression generally falls into two categories: (1) those that compress explicit 3-D geometric representations, like polygonal meshes with structured surface connectivity or point clouds capturing unstructured spatial data, and (2) those that operate directly on MVV frames with known camera poses. We focus on the latter. Traditional MVV compression has relied on explicit geometric modeling to reduce inter-view redundancy. Early binocular approaches used disparity estimation~\cite{vetro2011overview}, while 3D-HEVC~\cite{tech2015overview} integrated disparity prediction into 2-D codecs. More recently, MIV~\cite{boyce2021mpeg} consolidated multi-view data into a 2-D atlas, enabling compatibility with standard video encoders such as HEVC~\cite{sullivan2012overview}. 

Recently, inspired by advances in implicit neural representations (INRs), INR-based volumetric video compression methods~\cite{fujihashi2024fv,kwan2024immersive,shin2025neural} have been proposed, which exploit inter-view and inter-frame dependency by interpolating neural network (typically multilayer perceptron, i.e. MLP) parameters that characterize the entire multi-view volume, similar to NeRFs or GS methods as discussed above. Benefiting from this, the latest INR-based MVV compression methods~\cite{kwan2024immersive,zhu2025implicit} are associated with relatively low decoding complexity and have demonstrated superior coding performance to the MPEG test model for immersive video coding, TMIV~\cite{boyce2021miv}.

To support MVV research, many volumetric datasets have been created, and most of these are human-centric. For example, the 8iVFB voxelized point cloud~\cite{d2019jpeg} of full bodies and the MVUB upper body~\cite{loop2016microsoft} databases have been adopted in the Common Test Conditions of MPEG MIV. Other high-quality volumetric datasets include UVG-VPC~\cite{gautier2023uvg}, Vologram \& V-SENSE~\cite{zerman2020textured}, Owlii~\cite{xu2017owlii} and BVI-CR~\cite{gao2024bvi}. However, many of the aforementioned volumetric human datasets only provide the voxelized point clouds, without access to the corresponding raw multi-view RGB-D frames or detailed annotations such as body keypoints and foreground segmentation. The absence of this data prevents the use of mature, high-efficiency video coding infrastructures, which even in V-PCC standards~\cite{li2024mpeg} are fundamentally based on 2-D image or atlas representations. Furthermore, some of these datasets~\cite{zheng2024pku,jiang2024robust} can solely be used to evaluate the reconstruction quality rather than to measure rate-distortion (quality) performance as for the video compression task - this results in a disconnection from the actual volumetric video production pipelines.

\section{The \name \ Dataset}
\label{sec:dataset}

This section presents technical details on the raw source data, the content acquisition process, the generation of the supplementary data, and the information regarding accessibility. This project was approved by the University of Bristol Research Ethics Committee (Ref 21786), and all human participants have provided written consent.

\subsection{Content Coverage and Data Formats}\label{sec: source sequence}

\begin{figure}[t]
    \centering
    \begin{tikzpicture}
        \node[anchor=south west,inner sep=0] (image) at (0,0) {\includegraphics[width=1.\linewidth]{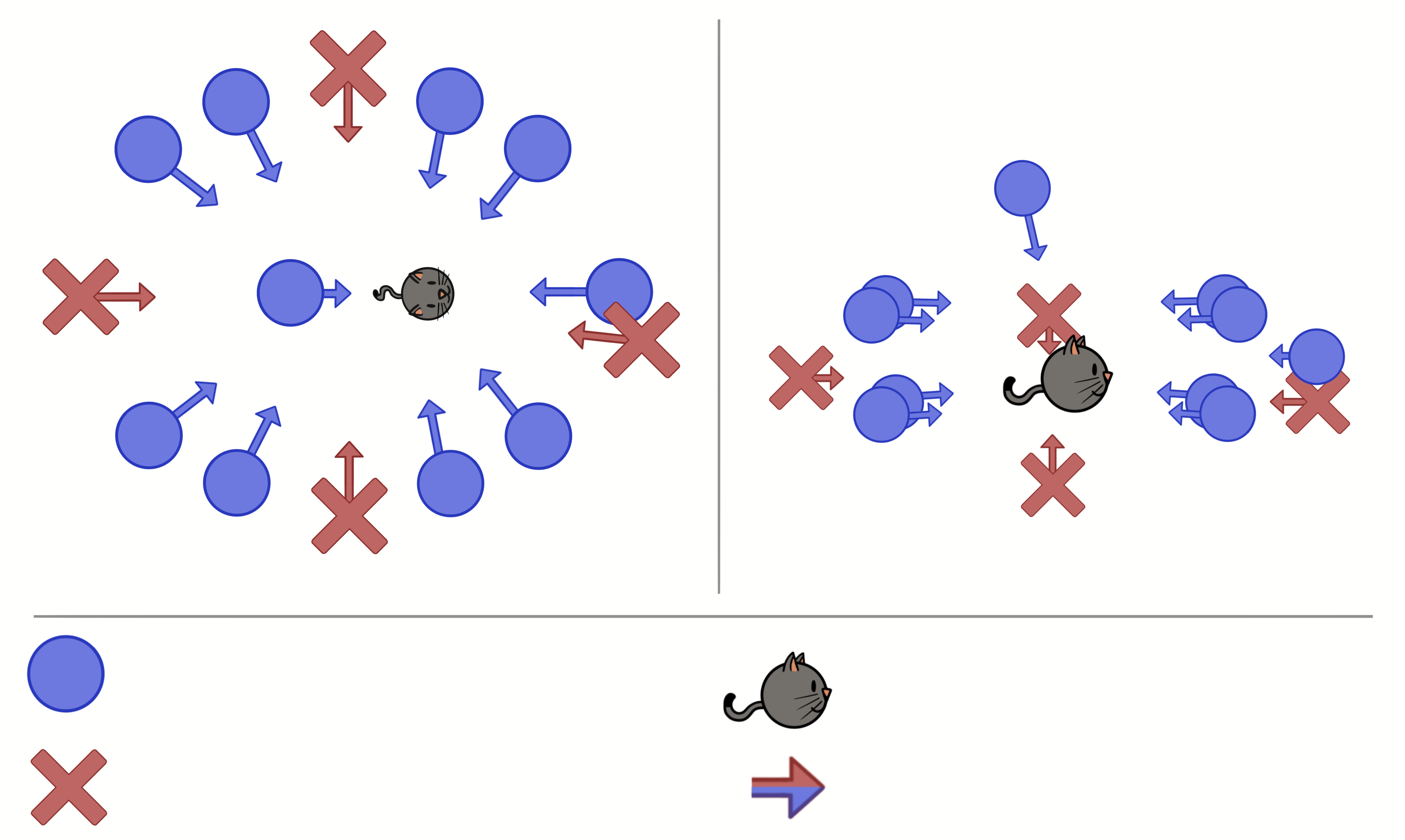}};

        \node at ($(image.south west) + (2.2,5.6)$) {\textcolor{black}{Top View}};
        \node at ($(image.south west) + (6.7,5.6)$) {\textcolor{black}{Left View}};

        \node at ($(image.south west) + (2.2,1.)$) {\textcolor{blue}{Training Camera}};
        \node at ($(image.south west) + (2.15,0.35)$) {\textcolor{red}{Testing Camera}};

        \node at ($(image.south west) + (6.9,1.)$) {\textcolor{black}{Human Performer}};
        \node at ($(image.south west) + (6.9,0.35)$) {\textcolor{black}{Direction}};

    \end{tikzpicture}    
    \caption{The camera placement is spherical around the human performer with a bias towards the performer's front side. This is shown in the \textit{Top View} (left) and \textit{Tilted Left View} (right) designs}
    \label{fig: camera placement}
\end{figure}
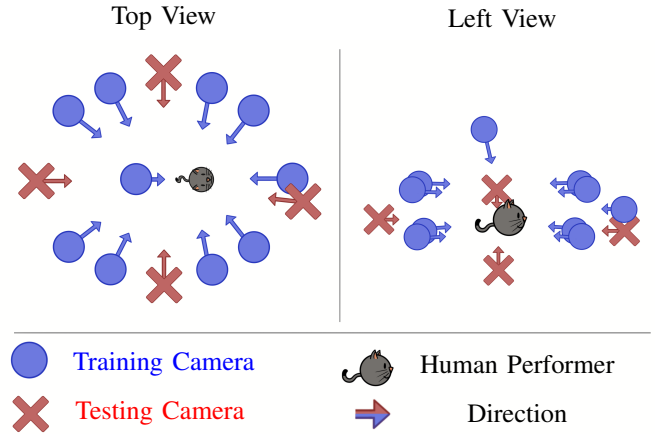

The \name \  dataset consists of thirty two unique scenes capturing human activities. This contains synchronized and calibrated RGB-D videos captured using 14 Azure Kinect cameras. Each RGB video has a spatial resolution of 2560$\times$1440 pixels at 30FPS, while the spatial and temporal resolutions of the depth video are 640$\times$576 pixels and 30FPS. For the same view, the RGB camera is positioned next to its corresponding depth camera, forming RGB-D pairs. The RGB-D camera pairs are placed spherically around the performer, as shown in \autoref{fig:mainfig} and \autoref{fig: camera placement}. In addition to RGB-D videos, we also provide associated metadata including camera poses, intrinsic parameters, and audio data. The poses consist of a full 6DoF orientation and translation, and the intrinsic parameters consist of focal lengths, principal points, and distortion coefficients. This metadata is provided per frame.

The 32 human-centric scenes involve 12 actors, 20 costumes and 35 dynamic objects. Eight scenes contain more than one person. Among all these actors, there are 6 European, 5 Asian and one African actor (and one dog), offering a relatively balanced racial diversity pool. This is also complemented by four special scenes: (i) with an actor in a full-body green suit, (ii) with an actor in an inflatable unicorn costume, (iii) with an actor in a clown wig, face paint and costume, (iv) with a multi-person multi-cultural scene consisting of one white European, one Asian and one African actor. Moreover, regarding the gender coverage, there are 11  scenes containing female performers, 23 with male performers, and three scenes include both male and female participants. Finally, 25\% of these scenes contain actors older than 50 years old, 25\% at the age between 30 and 50, and 51\% who are younger than 30.

Regarding the type of activities, 41\% of scenes are musical performances, 31\% are sport related and 28\% contain special actions. The musical performances include keyboard, bass, guitar, flute, cymbals and triangle solos, and also guitar duos with cymbals, triangle and flute. The sport sequences contain slow and fast paced pop and break dancing, tennis, weight lifting, stretching with elastic bands and a duo playing basketball. The solo special sequences include cosplay-posing, a clown playing with and popping shiny balloons, and various performers in banana costumes juggling or drinking colored liquids from transparent cups. The multi-person special sequences include duos in banana costumes doing similar activities, a trio singing happy birthday with cake and candles and a dog playing with its owner.

While many existing datasets claim a high diversity of people, hair and clothing, they lack other forms of visual diversity relating to video entertainment. \name \ overcomes this by capturing many sequences involving dynamic textures, dynamic transparent and reflective materials. To support even greater reconstruction and compression challenges, several of these scenes also include multiple people and objects. All these features are  summarized in \autoref{tab:large_dataset_comparison}, where \name \ is compared with other commonly used volumetric video datasets, in terms of capture information, formats, and content diversity. It is shown that among all these databases, only \name \ offers all different content types and information, which confirms its diversity and completeness.

\begin{table*}[t]
\caption{\textbf{Dataset comparisons:} \name \ is compared with the SotA on various characteristics and challenges. \textit{N/R} means ``not reported''. \textit{N/A} means ``not applicable''. ``\protect \rightdownarrow'' indicates additional RGB cameras.}
    \label{tab:large_dataset_comparison}
    \centering
    \resizebox{\linewidth}{!}{\begin{tabular}{r | c | c | c | c | c | c | c | c  }
    \toprule
     \textbf{Dataset} & \multicolumn{4}{c|}{\textbf{Capture Information}} & \multicolumn{2}{c|}{\textbf{RGB Data}} &  \multicolumn{2}{c}{\textbf{Depth Data}}\\ 
     \cmidrule{2-9}
     & \# Sequences & Length & Per-Frame Calib. & Layout & FPS & \# Views$\times$(Resolution)  & FPS & \# Views$\times$(Resolution)  \\ \midrule
        AIST/AIST++ \cite{tsuchida2019aist, li2021ai} & 1898 & 20-600s & \cellcolor{red!25}No & Circular & N/R & 3-9$\times$(1080P) & N/A & N/A \\
        CMU Panoptic \cite{joo2015panoptic} & 65 & 300-1200s & \cellcolor{red!25}No & Spherical & 25, & 480$\times$(480P), & 30, & 10$\times$(512$\times$424), \\
        \rightdownarrow &  &  &  &  & 30 & 31$\times$(1080P) &  \\

        DNA-Rendering \cite{cheng2023dna} & 1187 & 15s & \cellcolor{red!25}No & Cylindrical & 15 & 12$\times$(4K), 48$\times$(2K) & 15 & 8$\times$(N/R)\\
        DynaCap \cite{habermann2021real} & 5 & 40s & \cellcolor{red!25}No & Cylindrical & 25 & 120$\times$(4k) & N/A & N/A \\
        GeneBody1.0 \cite{cheng2022generalizable} & 370 & 10s & \cellcolor{red!25}No & Cylindrical & 20 & 48$\times$(N/R) & N/A & N/A \\
        Human3.6M  \cite{ionescu2013human3} & 17 & N/R & \cellcolor{red!25}No & Circular & 50 & 4$\times$(1000$\times$1000) & 25 & 1$\times$(176$\times$144)\\
        HUMBI \cite{yu2020humbi} & 772 & 5-6s & \cellcolor{red!25}No & Cylindrical &  60 & 107$\times$(1080P) & N/A & N/A \\
        HuMMan \cite{cai2022humman} & 339 & N/R & \cellcolor{red!25}No & Cylindrical & 30, &  12$\times$(4K), & 30, & 10$\times$(640$\times$576),\\
        \rightdownarrow &  & & &  & 60 &  10$\times$(1080P) & 30 & 12$\times$(256$\times$192)  \\

        MVHuamnNet \cite{xiong2024mvhumannet} & N/R & N/R & \cellcolor{red!25}No & Cylindrical & N/R & 48$\times$(4K), 24$\times$(2K) & N/A & N/A \\
        ZJU-MoCap  \cite{peng2021zju} & 9 & 3-14s & \cellcolor{red!25}No & Cylindrical & 21 & 21$\times$(N/R) & N/A & N/A \\
        BVI-CR \cite{gao2024bvi} & 18 & 16-20s & \cellcolor{red!25}No & Spherical & 25 & 10$\times$(1080P) & 25 & 10$\times$(512$\times$424) \\\midrule
       \textbf{\name \ (ours)}& 32 & 5-120s &\cellcolor{green!25} Yes & Spherical & 30 & 14$\times$(2K) & 30 & 14$\times$(640$\times$576) \\ \midrule\midrule
    \end{tabular}}

    \resizebox{\linewidth}{!}{\begin{tabular}{r | c|c|c|c|c|c |c|c|c|c|c }
       \textbf{Dataset} & \multicolumn{6}{c|}{\textbf{Diversity Challenges}} &\multicolumn{2}{c|}{\textbf{Multi-}} & \multicolumn{3}{c}{\textbf{Dynamic Materials}} \\ 
        \cmidrule{2-12}
       & Age & Sex & Skin & Hair & Clothing & Action Type & Person & Object & Dyn. Tex. & Transparent & Reflective \\\midrule
        AIST/AIST++ \cite{tsuchida2019aist, li2021ai}& \cellcolor{green!25}Yes & \cellcolor{green!25}Yes & \cellcolor{green!25}Yes & \cellcolor{green!25}Yes & \cellcolor{green!25}Yes & \cellcolor{red!25}No   & \cellcolor{green!25}Yes & \cellcolor{red!25}No & \cellcolor{red!25}No & \cellcolor{green!25}Yes & \cellcolor{green!25}Yes  \\
        CMU Panoptic \cite{joo2015panoptic} & \cellcolor{green!25}Yes & \cellcolor{green!25}Yes & \cellcolor{green!25}Yes & \cellcolor{green!25}Yes & \cellcolor{red!25}No & \cellcolor{green!25}Yes & \cellcolor{green!25}Yes & \cellcolor{green!25}Yes & \cellcolor{red!25}No & \cellcolor{red!25}No & \cellcolor{green!25}Yes \\
        DNA-Rendering \cite{cheng2023dna} & \cellcolor{green!25}Yes & \cellcolor{green!25}Yes & \cellcolor{green!25}Yes & \cellcolor{green!25}Yes & \cellcolor{green!25}Yes & \cellcolor{green!25}Yes  & \cellcolor{red!25}No & \cellcolor{green!25}Yes & \cellcolor{red!25}No & \cellcolor{green!25}Yes & \cellcolor{red!25}No  \\
        DynaCap \cite{habermann2021real} & \cellcolor{red!25}No & \cellcolor{red!25}No & \cellcolor{red!25}No & \cellcolor{red!25}No & \cellcolor{red!25}No & \cellcolor{red!25}No & \cellcolor{red!25}No &\cellcolor{red!25}No & \cellcolor{red!25}No & \cellcolor{red!25}No & \cellcolor{red!25}No\\
        GeneBody1.0 \cite{cheng2022generalizable} & \cellcolor{green!25}Yes &  \cellcolor{green!25}Yes & \cellcolor{green!25}Yes & \cellcolor{green!25}Yes & \cellcolor{green!25}Yes & \cellcolor{green!25}Yes & \cellcolor{red!25}No & \cellcolor{green!25}Yes & \cellcolor{red!25}No & \cellcolor{red!25}No & \cellcolor{green!25}Yes \\
        Human3.6M \cite{ionescu2013human3}&   \cellcolor{red!25}No & \cellcolor{green!25}Yes &   \cellcolor{red!25}No &  \cellcolor{red!25}No &   \cellcolor{red!25}No &  \cellcolor{red!25}No & \cellcolor{red!25}No & \cellcolor{green!25}Yes & \cellcolor{red!25}No & \cellcolor{red!25}No & \cellcolor{red!25}No \\
        HUMBI \cite{yu2020humbi} & \cellcolor{green!25}Yes & \cellcolor{green!25}Yes & \cellcolor{green!25}Yes & \cellcolor{green!25}Yes &  \cellcolor{red!25}No &   \cellcolor{red!25}No & \cellcolor{red!25}No &\cellcolor{red!25}No & \cellcolor{red!25}No & \cellcolor{red!25}No & \cellcolor{red!25}No\\
        HuMMan \cite{cai2022humman} &   \cellcolor{red!25}No & \cellcolor{green!25}Yes & \cellcolor{red!25}No &  \cellcolor{red!25}No &   \cellcolor{red!25}No &  \cellcolor{red!25}No & \cellcolor{red!25}No &\cellcolor{red!25}No & \cellcolor{red!25}No & \cellcolor{red!25}No & \cellcolor{red!25}No \\
        MVHuamnNet \cite{xiong2024mvhumannet} & \cellcolor{green!25}Yes & \cellcolor{green!25}Yes & \cellcolor{red!25}No & \cellcolor{red!25}No & \cellcolor{red!25}No & \cellcolor{red!25}No & \cellcolor{red!25}No &\cellcolor{red!25}No & \cellcolor{red!25}No & \cellcolor{red!25}No &  \cellcolor{green!25}Yes  \\
        ZJU-MoCap \cite{peng2021zju} & \cellcolor{red!25}No & \cellcolor{green!25}Yes & \cellcolor{red!25}No & \cellcolor{red!25}No & \cellcolor{red!25}No & \cellcolor{red!25}No & \cellcolor{red!25}No &\cellcolor{red!25}No & \cellcolor{red!25}No & \cellcolor{red!25}No & \cellcolor{red!25}No \\ 
        BVI-CR \cite{gao2024bvi} & \cellcolor{red!25}No & \cellcolor{red!25}No & \cellcolor{red!25}No & \cellcolor{red!25}No & \cellcolor{red!25}No & \cellcolor{green!25}Yes & \cellcolor{green!25}Yes &\cellcolor{red!25}No & \cellcolor{red!25}No & \cellcolor{red!25}No & \cellcolor{red!25}No \\  \midrule
       \textbf{\name \ (ours)}  & \cellcolor{green!25}Yes & \cellcolor{green!25}Yes & \cellcolor{green!25}Yes & \cellcolor{green!25}Yes & \cellcolor{green!25}Yes & \cellcolor{green!25}Yes& \cellcolor{green!25}Yes & \cellcolor{green!25}Yes & \cellcolor{green!25}Yes & \cellcolor{green!25}Yes & \cellcolor{green!25}Yes  \\ \bottomrule
    \end{tabular}}
\end{table*}

\subsection{Content Acquisition}\label{subsec: additional data}
\begin{figure}
    \centering
    \includegraphics[width=0.49\linewidth]{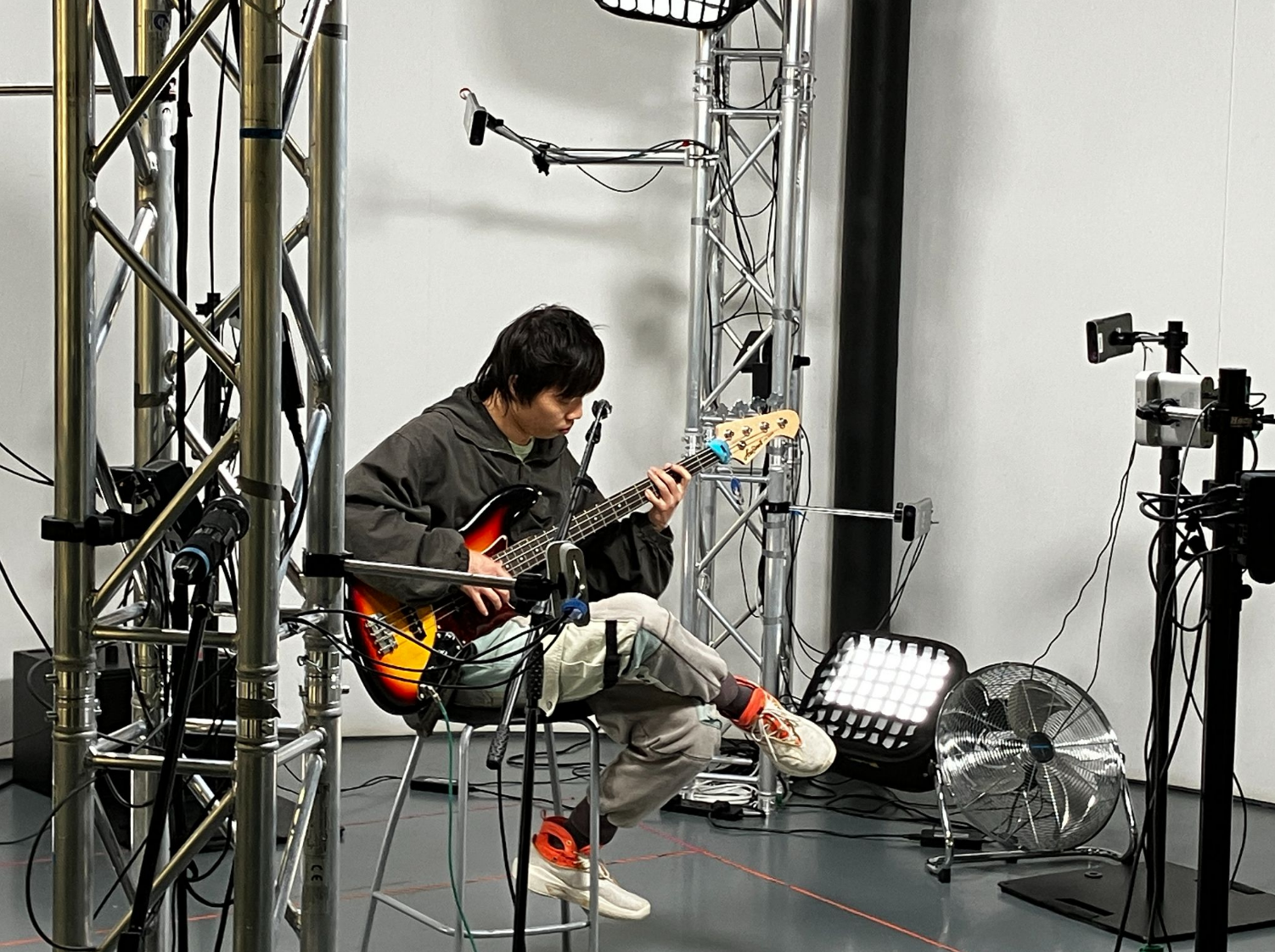}
    \hfill
    \includegraphics[width=0.49\linewidth]{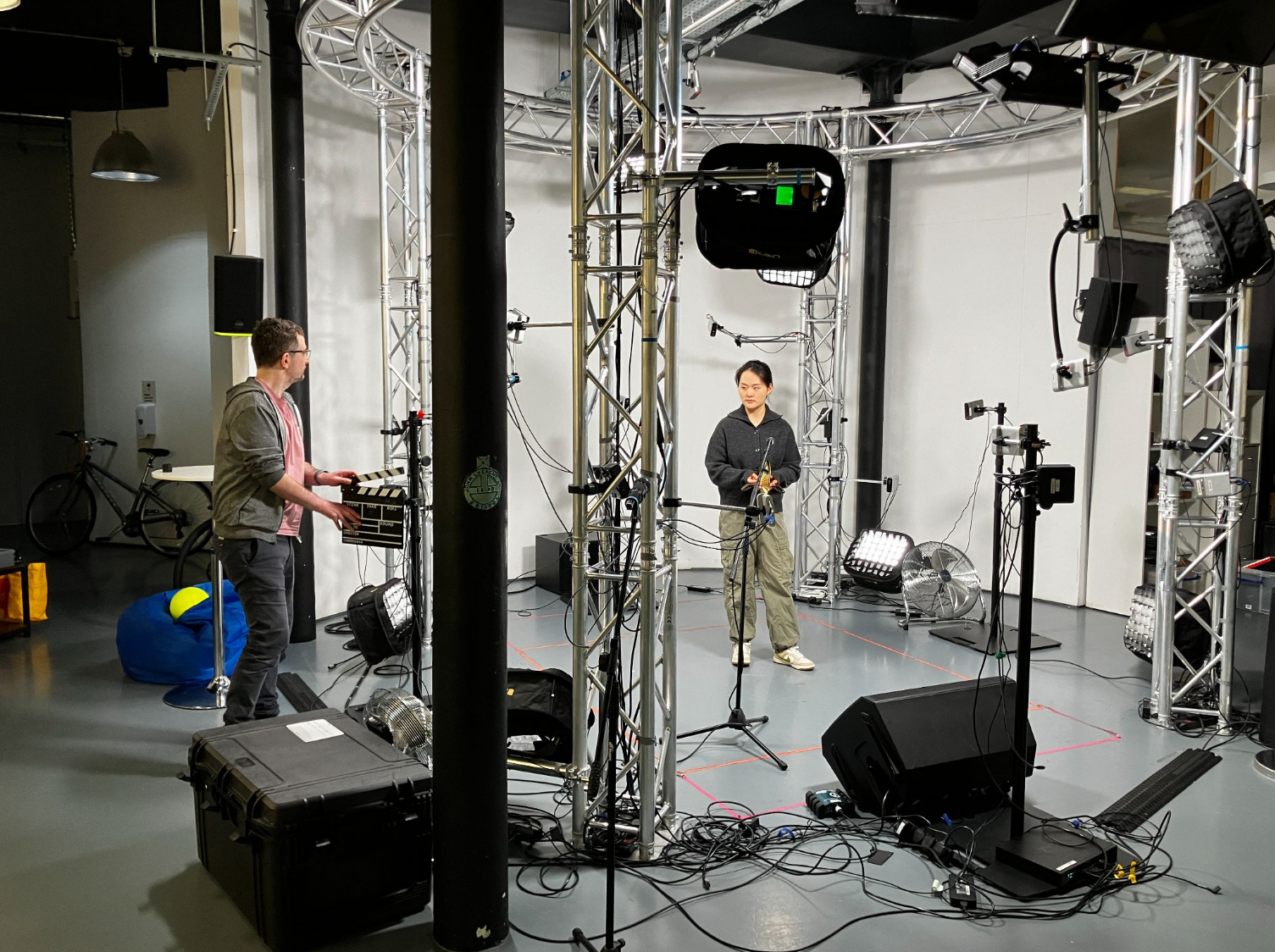}
    \caption{Inside The Metaverse Studio: The truss columns and training cameras attached to arms, as well as the testing cameras on stands can be seen. In the right image a staff member is directing a performer}
    \label{fig:inside_the_studio}
\end{figure}

All the raw data were captured at the Metaverse Studio, a live 3-D virtual production stage hosted and managed by Condense Reality\footnote{\url{https://condense.live/}}; shown in \autoref{fig:inside_the_studio}. The studio has been used for live volumetric broadcasting events\footnote{``Live'' translates to ``Vivo'' in Italian.}, where MVVs are used to generate and stream dynamic volumetric mesh representations of performers for virtual concerts. The staff involved in the data collection consists of an audio and a camera engineer who are responsible for capture and calibration. The calibration was accomplished using Condense Reality's proprietary method. As the system is designed for  live, high-quality volumetric production and delivery,  the calibration is accurate and reliable.

\begin{figure}[t]
    \centering
    \begin{tikzpicture}
        \node[anchor=south west,inner sep=0] (image) at (0,0) {\includegraphics[width=1.\linewidth]{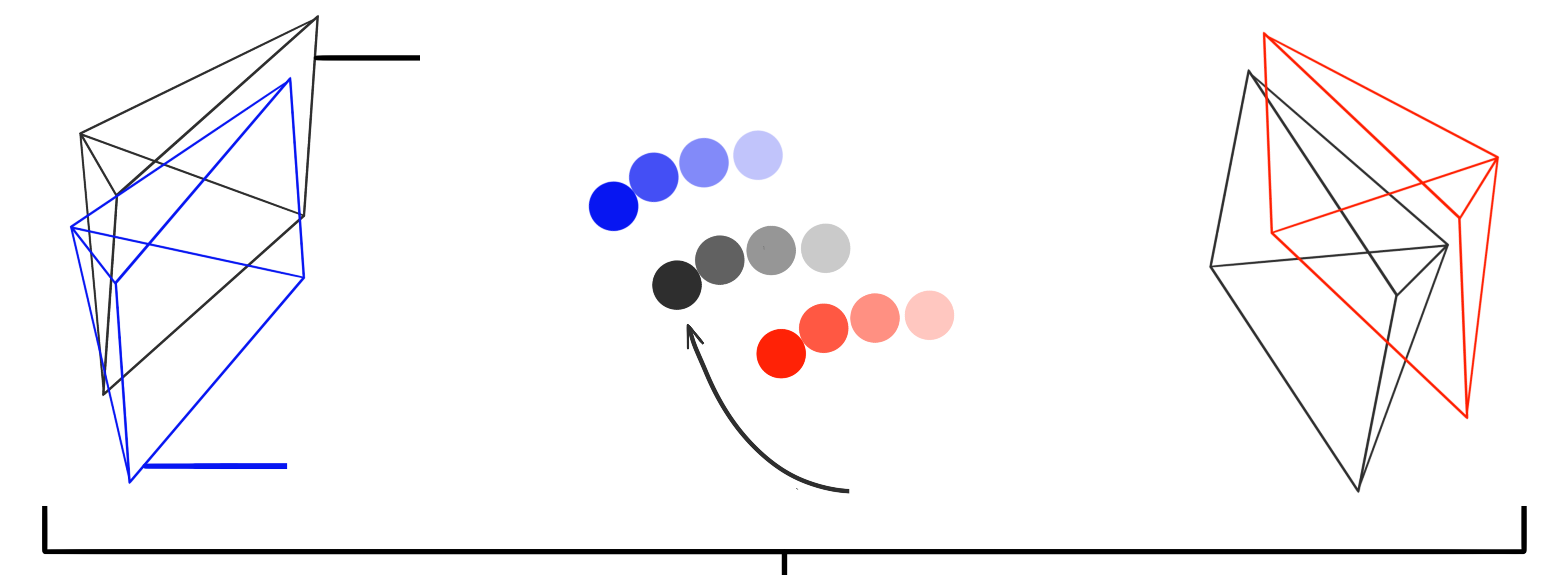}};
        
        \node[below=0.5em of image, align=center] {
            \textbf{BundleAdjustment}[\textcolor{blue}{Blue}, \textcolor{red}{Red}, ...] $\rightarrow$ \textcolor{black}{Black}
        };

        \node at ($(image.south west) + (2.9,0.6)$) {\textcolor{blue}{Un-aligned Pose}};
        \node at ($(image.south west) + (3.4,2.9)$) {\textcolor{black}{Aligned Pose}};
        \node at ($(image.south west) + (6.2,0.45)$) {\textcolor{black}{Aligned Trajectory}};

    \end{tikzpicture}    
    \caption{Before filming, bundle adjustment is used on camera groups to optimize camera poses by aligning the known trajectory of a segmented object}
    \label{fig: calibration example}
\end{figure}

During calibration, a 3-D marker was moved within the field-of-view of all the cameras. For each camera, the marker was segmented and a candidate 3-D position marker was estimated. Over time, the marker traced a trajectory in 3-D space. This is aligned with trajectories from the other cameras using a proprietary bundle adjustment strategy, as shown in \autoref{fig: calibration example}. The intrinsic parameters were initialized with the factory settings and optimized by minimizing the difference between the projection of the 3-D marker via the intrinsics and the segmentation of the real marker in the images captured during calibration.

\subsection{Additional Data} \label{sec: additional-data}

\begin{figure}
    \centering
    \begin{tikzpicture}
        \node[anchor=south west,inner sep=0] (image) at (0,0) {\includegraphics[width=1.\linewidth]{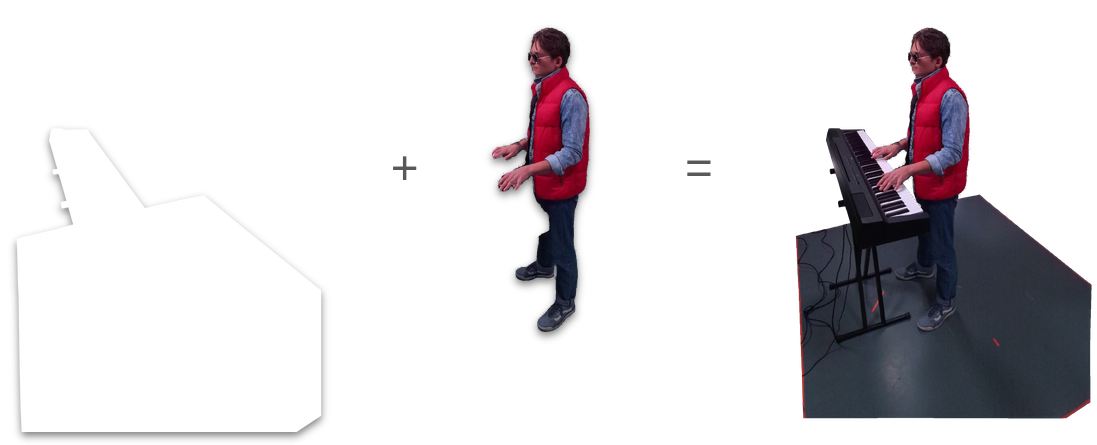}};
        
        \node at ($(image.south west) + (1.3,-0.4)$) {\textcolor{black}{Static Mask}};
        \node at ($(image.south west) + (4.5,-0.4)$) {\textcolor{black}{Dynamic Mask}};
        \node at ($(image.south west) + (7.8,-0.4)$) {\textcolor{black}{Scene Mask}};

    \end{tikzpicture}
    \caption{Adding static masks (left) and dynamic SAM2 masks (middle) to produce a scene mask (right)}
    \label{fig: 2dmasks}
\end{figure}
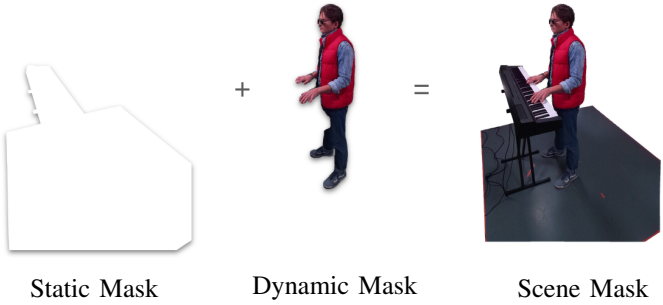

In addition to the raw calibrated data, we also provide GUIs that facilitate the process of generating additional data. The additional data includes 2-D dynamic foreground masks, colored point clouds, and 3-D depth filters and box masks. Here the 2-D masks are constructed by adding a dynamic foreground mask with a manually drawn static mask, shown in \autoref{fig: 2dmasks}. The dynamic foreground mask is generated via a GUI that implements SAM2 for our dataset. The static masks are manually drawn for each camera and are the same for all frames. The static masks comprise the main stage area outlined by red tape (see \autoref{fig: 2dmasks}). Summing the two masks segments the objects inside the stage area while minimizing the segmentation errors due to high-contrast shadows (mostly present around the feet). The GUI also provides mask dilation and erosion functionality.

A 3-D colored point cloud is obtained with the following steps. For each RGB-D camera pair, the pixel values from the depth camera are scaled and projected into world space as a point cloud, $X$ via $M_d$ derived from the known intrinsic and extrinsic camera parameters for a given frame. Subsequently, the point cloud $X$ is projected into the RGB camera space, with $M_{c}^{-1}$. Using the RGB image, $I_{RGB}$, the color of each pixel, $c_{x,y,z}$ positioned at $(u,v,0)$, is assigned to the nearest point along the z-axis, in front of the camera. This whole process is described by Equations 1-5, where $x_d$ and $x_c$ are the point cloud positions in the depth and color camera spaces, respectively, and $z_k$ is the depth of each point, $k$, along the z-axis corresponding to the pixel $x_{u,v,0}$.
\begin{align}
    X &= M_d x_d, \\
    x_c &= M_{c}^{-1} X, \\
    c_{u,v,0} &\rightarrow x_{(u,v,z)},\\
    \quad \text{where} \quad c_{u,v,0} &\in I_{RGB}, \\ 
    \text{and} \quad x_{u,v,z} &= \arg \min_{k} \left( z_k > 0\right).
\end{align}
With 10 training RGB-D camera pairs, a total of $3.7$ million points are produced per frame - this is equivalent to $111$ million points per second. This increases to $155$ million points per second if all 14 camera pairs are used. Due to potential computational constraints in practical use, for each sequence we provide (i) a depth filter (ii) a 3-D box mask, and (iii) a downsampling tool that uses off-the-shelf voxel grid downsampling \cite{zhou2018open3d}, to cull points outside and inside the box mask separately. The 3-D box mask is predefined and does not require generation or labeling.

\subsection{Accessibility}\label{sec: access}

This open-access dataset is accompanied by two GUIs. The raw data, source code and documentation are all  available online. Additionally, all scenes have been cataloged online. This allows users the ability to cut sequences of varying lengths w.r.t their needs. Finally, the subset of data used in the following experiments is also made publicly available.

\section{Experiment Design}\label{sec: experiment design}
To demonstrate the use of the database for dynamic NVS and MVV compression tasks, benchmark experiments have been designed involving SotA reconstruction and compression methods. To ensure a fair volumetric evaluation, for each scene, we select four testing cameras which are situated in front, behind, to the left, and to the right of the capture area, meaning the other 10 cameras are used for trainring. Both tasks were conducted on an NVidia RTX 4090 with 24GB of vRAM.

\subsection{Dynamic Novel View Synthesis}
We evaluate \name \  as a 6DoF sparse-view challenge by benchmarking three SotA dynamic 3-D reconstruction methods, 4D-GS \cite{wu20244d}, STG \cite{li2024spacetime} and SC-GS \cite{huang2024sc}, on six selected scenes, each lasting 300 frames. The dynamic NVS \name \  subset and related code are available online to ensure reproducibility. The main challenge with this task is that test views will contain irrelevant background details that are not seen by the training views. To overcome this, we train each model on the full images but test using masked images. We do not train with masks as some scenes contain dynamic transparent and reflective materials, which are nontrivial to segment. Furthermore, the test masks are generated using a dilation kernel that is larger than the erosion kernel. This smoothens and pads the edges of the foreground so that fine edge details (e.g. hair) and transparent materials (e.g. glass) are fully captured during evaluation.

Another challenge relates to the time input associated with each frame. For most existing datasets, the timestamp is calculated based on the normalized frame position, where \[\hat{t} = \text{frame index} / (N-1)\] for $N$ frames, leading to a minimum precision of $3.3ms$ for 300 frames. While this is not currently a major problem, it may become an issue in the future for scenes containing: (1) fast motion, or (2) thin/fine dynamic objects. Therefore, as \name \ provides UTC timestamps per-camera and per-frame, we use raw normalized timestamps, \[\hat{t} = \frac{t - t_{min}}{t_{max} - t_{min}} \] where $t_{min}$ and $t_{max}$ are the earliest and latest UTC date-times w.r.t. all cameras in the dataset. This makes the precision of our dataset $1 \mu s$, which covers potential challenges in future investigations.

\begin{table*}[t]
    \caption{Objective dynamic NVS results on SotA dynamic volumetric reconstruction methods. Red indicates the best performance per scene and metric. * indicates that a model failed to generate motion}
    \label{tab: NVS results}
    \centering
    \resizebox{\linewidth}{!}{\begin{tabular}{c|r|cc|cc|cc|cc|c}
    \toprule
        \multirow{2}{*}{Scene} &\multirow{2}{*}{Model} &\multicolumn{2}{c|}{PSNR (dB)} & \multicolumn{2}{c|}{SSIM} & \multicolumn{2}{c|}{LPIPS-Alex} & \multicolumn{2}{c|}{LPIPS-VGG} & Model Size (MB) \\ 
        \cmidrule{3-10}
       & & Full & Mask& Full & Mask& Full & Mask& Full & Mask \\ \midrule 
      
       \multirow{3}{*}{Bassist}& 4D-GS \cite{wu20244d} &13.68& 19.82 & 0.5673 & 0.8837 & 0.4198 & 0.1342 & 0.4353 & 0.1471 & 135 \\
       & STG \cite{li2024spacetime}                    &\cellcolor{red!30}14.71 &\cellcolor{red!30}23.38 &\cellcolor{red!30}0.6447 &\cellcolor{red!30}0.9214 & \cellcolor{red!30}0.4136 & \cellcolor{red!30}0.1256 &\cellcolor{red!30} 0.4279 & \cellcolor{red!30}0.1256 & 249\\
       & *SC-GS \cite{huang2024sc}                     &13.46 & 20.00 & 0.3848 & 0.8314 & 0.5227 & 0.1776 & 0.5383 & 0.1904 & 338 \\ \midrule
       
       \multirow{3}{*}{Pianist}& 4D-GS \cite{wu20244d} &15.90&22.97&0.6143&0.9000&\cellcolor{red!30}0.3443&\cellcolor{red!30}0.1059&\cellcolor{red!30}0.4028&\cellcolor{red!30}0.1375& 138\\
       & STG \cite{li2024spacetime}                    &15.29&\cellcolor{red!30}23.61&\cellcolor{red!30}0.6239&\cellcolor{red!30}0.9143&0.3737&0.1148&0.4063&0.1427& 276 \\
       & *SC-GS \cite{huang2024sc}                     &\cellcolor{red!30}15.92&23.00&0.4574&0.8604&0.4819&0.1651&0.5117&0.1787& 262 \\ \midrule
       
       \multirow{3}{*}{Cymbals}& 4D-GS \cite{wu20244d} &13.65&20.30&0.5910&0.9219&0.4000&0.1029&0.4264&0.1296& 135 \\
       & STG \cite{li2024spacetime}                    &\cellcolor{red!30}14.05&\cellcolor{red!30}23.42&\cellcolor{red!30}0.6534&\cellcolor{red!30}0.9481&\cellcolor{red!30}0.3960&\cellcolor{red!30}0.0888&\cellcolor{red!30}0.4231&\cellcolor{red!30}0.1259& 293 \\
       & *SC-GS \cite{huang2024sc}                     &13.21&19.78&0.3994&0.8662&0.5271&0.1685&0.5411&0.1828 & 332 \\ \midrule
       
       \multirow{3}{*}{Curling}& 4D-GS \cite{wu20244d} &12.37&19.51&0.5302&0.9151&0.3988&0.1015&0.4440&\cellcolor{red!30}0.1310 & 133\\
       & STG \cite{li2024spacetime}                    &\cellcolor{red!30}13.09&\cellcolor{red!30}19.84&\cellcolor{red!30}0.5936&\cellcolor{red!30}0.9301&\cellcolor{red!30}0.3675&\cellcolor{red!30}0.0997&\cellcolor{red!30}0.4349&0.1376 & 283 \\
       & *SC-GS \cite{huang2024sc}                     &12.21&19.30&0.3975&0.8749&0.5032&0.1511&0.5339&0.1731 & 493 \\ \midrule
       
       \multirow{3}{*}{Fruit}& 4D-GS \cite{wu20244d}   &\cellcolor{red!30}16.40&24.04&0.6369&0.9264&\cellcolor{red!30}0.3278&0.0836&\cellcolor{red!30}0.3993&0.1155& 133 \\
       & STG \cite{li2024spacetime}                    &15.80&\cellcolor{red!30}26.84&\cellcolor{red!30}0.6806&\cellcolor{red!30}0.9570&0.3375&\cellcolor{red!30}0.0674&0.4016&\cellcolor{red!30}0.1080& 272  \\
       & *SC-GS \cite{huang2024sc}                     &16.07&24.15&0.4867&0.8984&0.4731&0.1414&0.5131&0.1588& 193\\ \midrule
       
       \multirow{3}{*}{Pony}& 4D-GS \cite{wu20244d}    &11.85&17.31&0.5795&0.9018&0.4131&\cellcolor{red!30}0.1532&0.4551&\cellcolor{red!30}0.1888& 133 \\
       & STG \cite{li2024spacetime}                    &\cellcolor{red!30}12.72&\cellcolor{red!30}17.54&\cellcolor{red!30}0.6265&\cellcolor{red!30}0.9104&\cellcolor{red!30}0.3906&0.1600&\cellcolor{red!30}0.4492&0.2000& 320\\
       & SC-GS \cite{huang2024sc}                      &11.40&16.38&0.3954&0.8027&0.5428&0.2281&0.5631&0.2519& 314 \\ \midrule 
       
       \multirow{3}{*}{Average}& 4D-GS \cite{wu20244d} &13.98&20.66&0.5865&0.9082&0.3840&	0.1136&	0.4272&	0.1416 & 135 \\
       & STG \cite{li2024spacetime}                    &14.28&22.44&0.6371&0.9302&0.3798&0.1094&0.4238&0.1400 & 282\\
       & SC-GS \cite{huang2024sc}                      &13.71&20.44&0.4202&0.8557&0.5085&0.1720&0.5335&0.1893&322 \\ \bottomrule
    \end{tabular}}
\end{table*}

To ensure a fair comparison, all models are initialized with the same colored point cloud. Using the methods discussed in \autoref{subsec: additional data}, the point cloud data was generated using the first frame from the RGB-D training cameras, and downsampled to less than 400K points. This is the expected number of points that all three models operate with. Finally, to objectively assess quality, we use the PSNR, SSIM \cite{wang2004ssim}, LPIPS-VGG \cite{zhang2018unreasonable} and LPIPS-Alex \cite{zhang2018unreasonable} metrics. We selected these as they are widely used for evaluating NVS experiments. For each metric, we provide the full and masked results. We do this to demonstrate the effectiveness of using masks for evaluating sparse-view datasets.

\subsection{Multi-view Video Compression}
In addition to the NVS task, we evaluated the performance of MVV compression models on the proposed dataset, following the practice in \cite{gao2024bvi}. Specifically, we selected three representative scenes for benchmarking, including `Curling', `Pianist', and `Pony'. We evaluated three compression algorithms including the test model for the latest MPEG multi-view videos standard MPEG MIV \cite{boyce2021miv}, TMIV 17.0 \cite{dziembowski2023common} (with VVenC \cite{vvenc} as the video codec), and two open-sourced INR-based MVV codecs MV-HiNeRV \cite{kwan2024immersive} and MV-IERV \cite{zhu2025implicit}. All these codecs perform encoding and decoding with the source video views (the same ten views for the dynamic NVS task as mentioned above), and the TMIV view synthesizer \cite{dziembowski2023common} is used for rendering four pose trace views. To ensure that only useful content (i.e. the performer) is being encoded, all of the above coding steps (including the online optimization steps for the learned codecs) were performed with per-view foreground masking applied to the multi-view frames and loss excluded for the masked background during both encoding and evaluation. For all codecs evaluated, we use TMIV to encode the depth as geometry and mask as occupancy maps. We align depth to the corresponding RGB cameras by first gathering the un-projected depth points to XYZ coordinates in the 3-D world space over all camera views, for the first 100 frames of each scene, then re-projecting these XYZ coordinates back to the RGB camera planes.

We follow the TMIV CTC \cite{dziembowski2023common}, encoding the first 65 frames with an \texttt{IntraPeriod} of 32. MV-HiNeRV and MV-IERV codecs are trained to minimize $l_1$-based distortion loss on the masked frames, where MV-HiNeRV simultaneously optimizes the rate loss in an end-to-end manner and MV-IERV adopts pruning and quantization without entropy penalization. After decoding using MV-HiNeRV and MV-IERV, the outputs were converted back to the original YUV 4:2:0 format, and we apply the same view synthesizer in TMIV to generate both source and pose trace views for evaluation. The Bj{\o}ntegaard Delta Rate (BD-rate)~\cite{BD} is used to measure the relative coding efficiency between codecs. The distortions are measured using PSNR, SSIM, and IV-PSNR~\cite{dziembowski2022iv} with regard to the uncompressed reference~\cite{dziembowski2023common}.

\section{Benchmarks}\label{sec: benchmarks}
\subsection{Dynamic Novel View Synthesis}
\begin{figure*}[t]
    \centering
    \includegraphics[width=1.\textwidth]{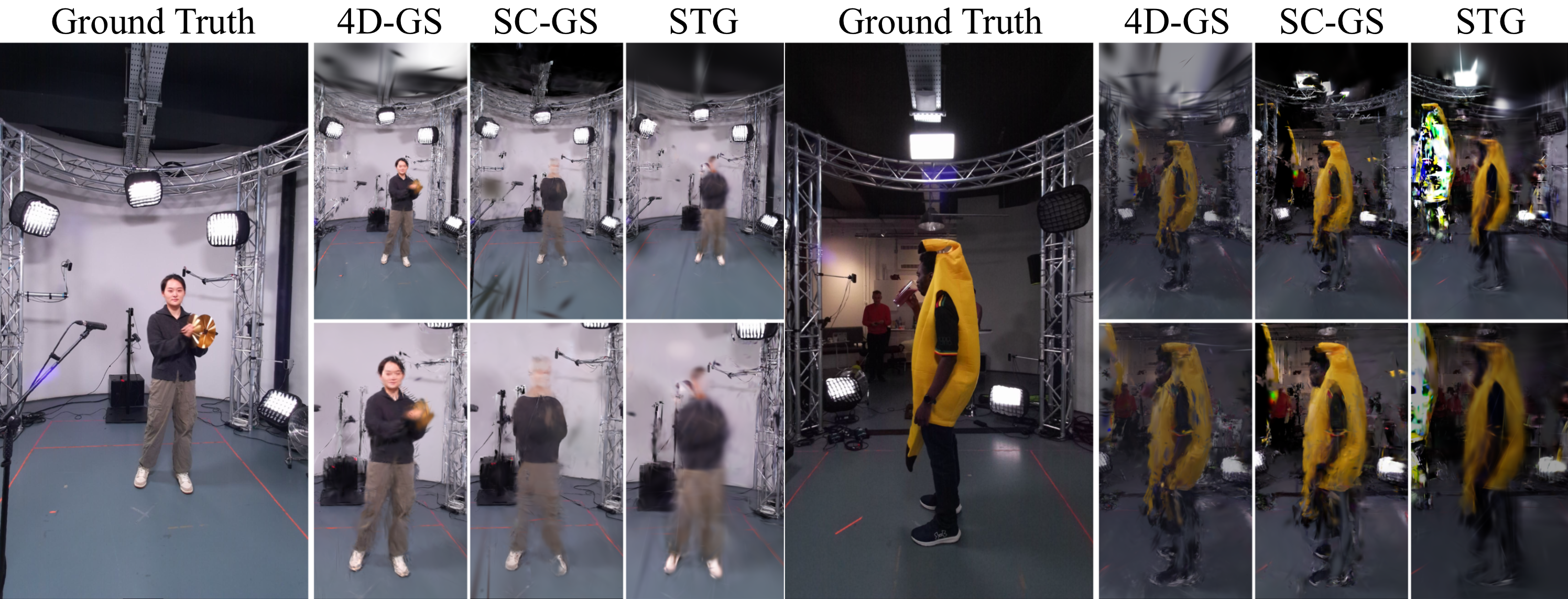}
    \caption{Visual comparison of dynamic reconstruction methods for each scene. The top row of each result is the full render and the bottom row is a zoomed image.}
    \label{fig: dynamic NVS visual results}
\end{figure*}

\begin{figure*}[t]
    \centering
    \includegraphics[width=1.\textwidth]{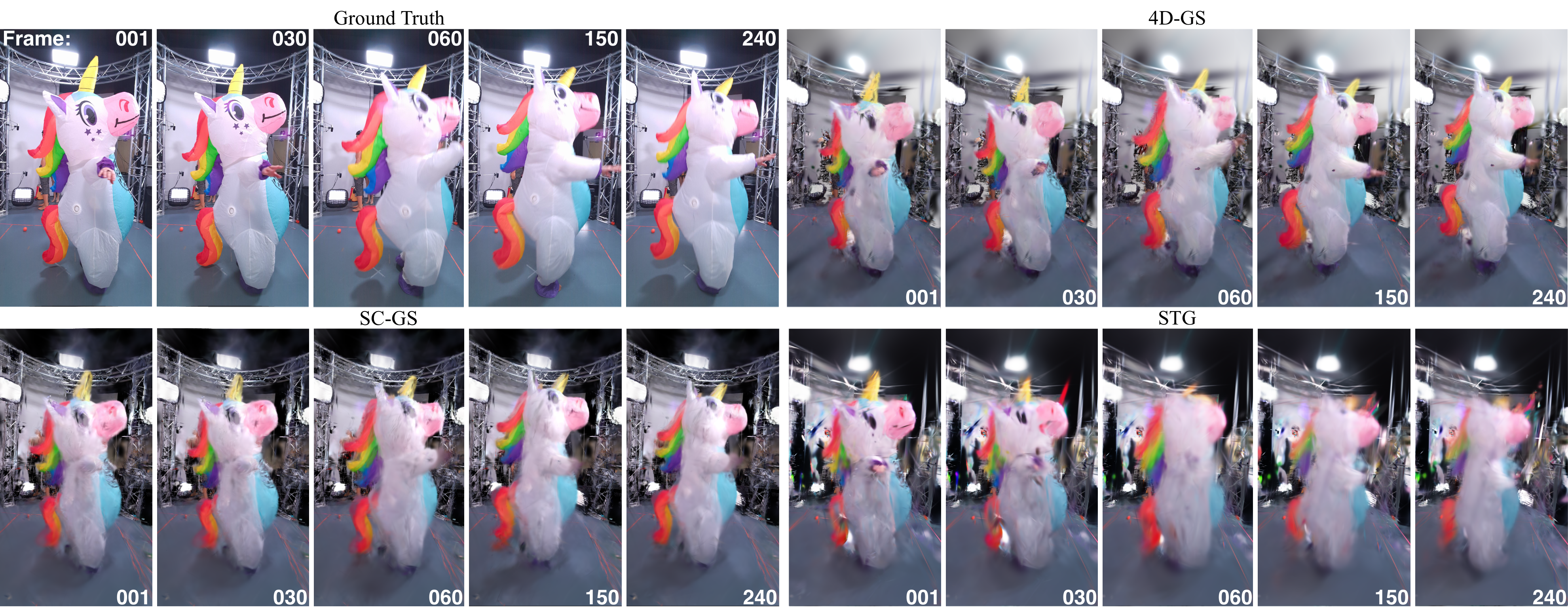}
    \caption{Visual comparison of NVS over time for evaluating temporal degradation. The Pony scene was selected to demonstrate: (1) STG degenerating with time, and (2) the quality of SC-GS when it successfully generates control points}
    \label{fig: dynamic NVS temporal results}
\end{figure*}

\autoref{tab: NVS results} presents the objective metric and \autoref{fig: dynamic NVS visual results} and \ref{fig: dynamic NVS temporal results} present the visual comparisons. The speed for each model is around 80FPS for 4D-GS, 200FPS for SC-GS and 140FPS for STG.
Validation and further video comparisons are presented online, demonstrating that 4D-GS and STG are capable of achieving full dynamic 3-D reconstruction, whereas SC-GS produces a high quality 3-D model but is not capable of driving motion for all scenes except the Pony scene. 


\begin{table*}[t]
\centering
\caption{Multiview video compression results for three test codecs. BD-Rate (\%) is reported separately for source and pose trace views, measured in PSNR, SSIM, and IV-PSNR, with TMIV as the anchor. We highlight the \colorbox{red!30}{best} results. We also provide complexity comparison including model size, MACs/pixel, encoding and decoding FPS. All results are measured on a 16-core Xeon(R) Gold 6430 CPU and an NVIDIA RTX 4090. FPS values for NeRV models are averaged across sequences and reported along with the range, as they are dependent on the model size. `N/A' denotes \textit{not applicable} and `A/A' denotes \textit{as above}.}
\label{tab:compression-results}
\resizebox{\linewidth}{!}{
\begin{tabular}{r|r|rr|rr|rr|ccc}
\toprule[1.2pt]
& Metric & \multicolumn{2}{c|}{PSNR} & \multicolumn{2}{c|}{SSIM} & \multicolumn{2}{c|}{IV-PSNR} & \multicolumn{3}{c}{Model complexity} \\
\cmidrule{1-8} \cmidrule{9-11}
Scene & Codec & Source & \makecell{Pose \\ trace} & Source & \makecell{Pose \\ trace} & Source & \makecell{Pose \\ trace} & Size (M) $\downarrow$ & kMACs/px $\downarrow$ & \makecell{Dec.\\FPS $\uparrow$} \\
\midrule
\multirow{3}{*}{Curling}
    & TMIV \cite{dziembowski2023common}   & 0.00 & 0.00 & 0.00 & 0.00 & 0.00 & 0.00 & N/A & N/A & 5.71 \\
    & MV-HiNeRV \cite{kwan2024immersive}   & \cellcolor{red!30}-38.01 & \cellcolor{red!30}-37.13 & \cellcolor{red!30}-40.17 & \cellcolor{red!30}-39.21 & \cellcolor{red!30}-39.59 & \cellcolor{red!30}-38.98 & $\text{8.42} \pm \text{6.58}$ & $\text{158.5} \pm \text{122.1}$ & $\text{29.8} \pm \text{8.71}$ \\
    & MV-IERV \cite{zhu2025implicit}       & -20.23 & -19.99 & -20.24 & -19.81 & -21.23 & -21.20 & $\text{8.21} \pm \text{6.38}$ & $\text{112.2} \pm \text{96.52}$ & $\text{48.6} \pm \text{13.3}$ \\
\midrule
\multirow{3}{*}{Pianist}
    & TMIV \cite{dziembowski2023common}    & 0.00 & 0.00 & 0.00 & 0.00 & 0.00 & 0.00 & N/A & N/A & 5.75 \\
    & MV-HiNeRV \cite{kwan2024immersive}   & \cellcolor{red!30}-46.63 & \cellcolor{red!30}-43.59 & \cellcolor{red!30}-44.88 & \cellcolor{red!30}-45.27 & \cellcolor{red!30}-48.02 & \cellcolor{red!30}-46.48 & A/A & A/A & A/A \\
    & MV-IERV \cite{zhu2025implicit}       & -14.69 & -14.26 & -13.35 & -13.21 & -13.80 & -12.79 & A/A & A/A & A/A \\
\midrule
\multirow{3}{*}{Pony}
    & TMIV \cite{dziembowski2023common}    & 0.00 & 0.00 & 0.00 & 0.00 & 0.00 & 0.00 & N/A & N/A & 5.75 \\
    & MV-HiNeRV \cite{kwan2024immersive}   & \cellcolor{red!30}-37.01 & \cellcolor{red!30}-36.98 & \cellcolor{red!30}-36.77 & \cellcolor{red!30}-36.70 & \cellcolor{red!30}-33.81 & \cellcolor{red!30}-34.99 & A/A & A/A & A/A \\
    & MV-IERV \cite{zhu2025implicit}       & -19.01 & -18.99 & -18.86 & -18.81 & -17.57 & -18.43 & A/A & A/A & A/A \\
\bottomrule[1.3pt]
\end{tabular}}
\end{table*}

\begin{figure*}[t]
    \centering
    \includegraphics[width=\linewidth]{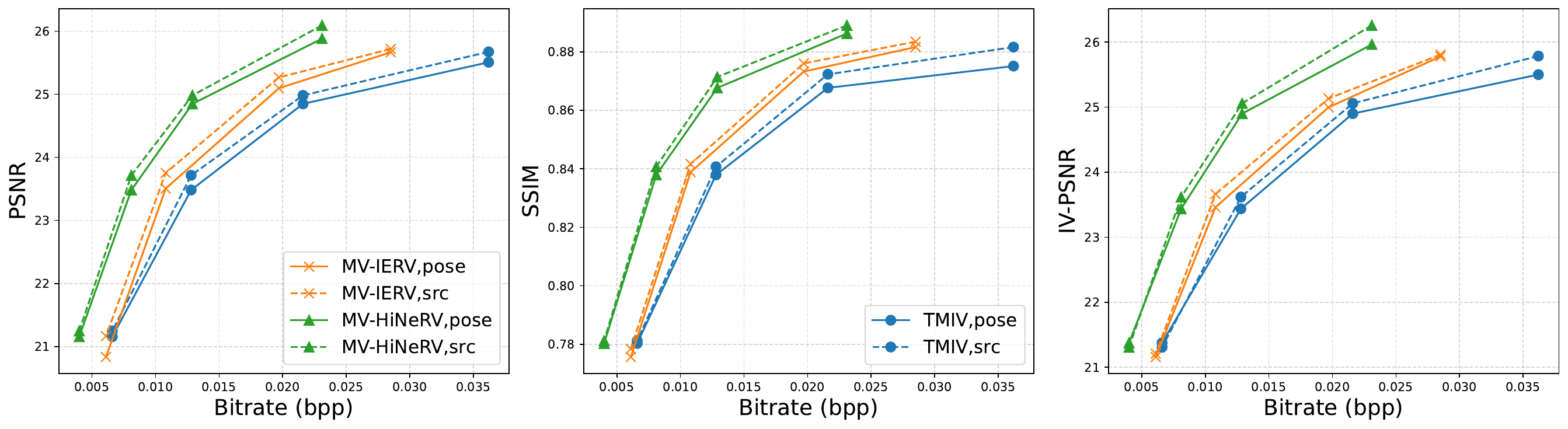}
    \caption{Rate-distortion performance comparison of selected baselines, with distortion measured by PSNR, SSIM, and IV-PSNR, respectively. Here, \textbf{src} denotes source views and \textbf{pose} denotes pose trace views.}
    \label{fig:rd-plot}
\end{figure*}

\begin{figure}[ht]
    \centering

    \begin{minipage}{\linewidth}
        \centering
        \includegraphics[width=\linewidth]{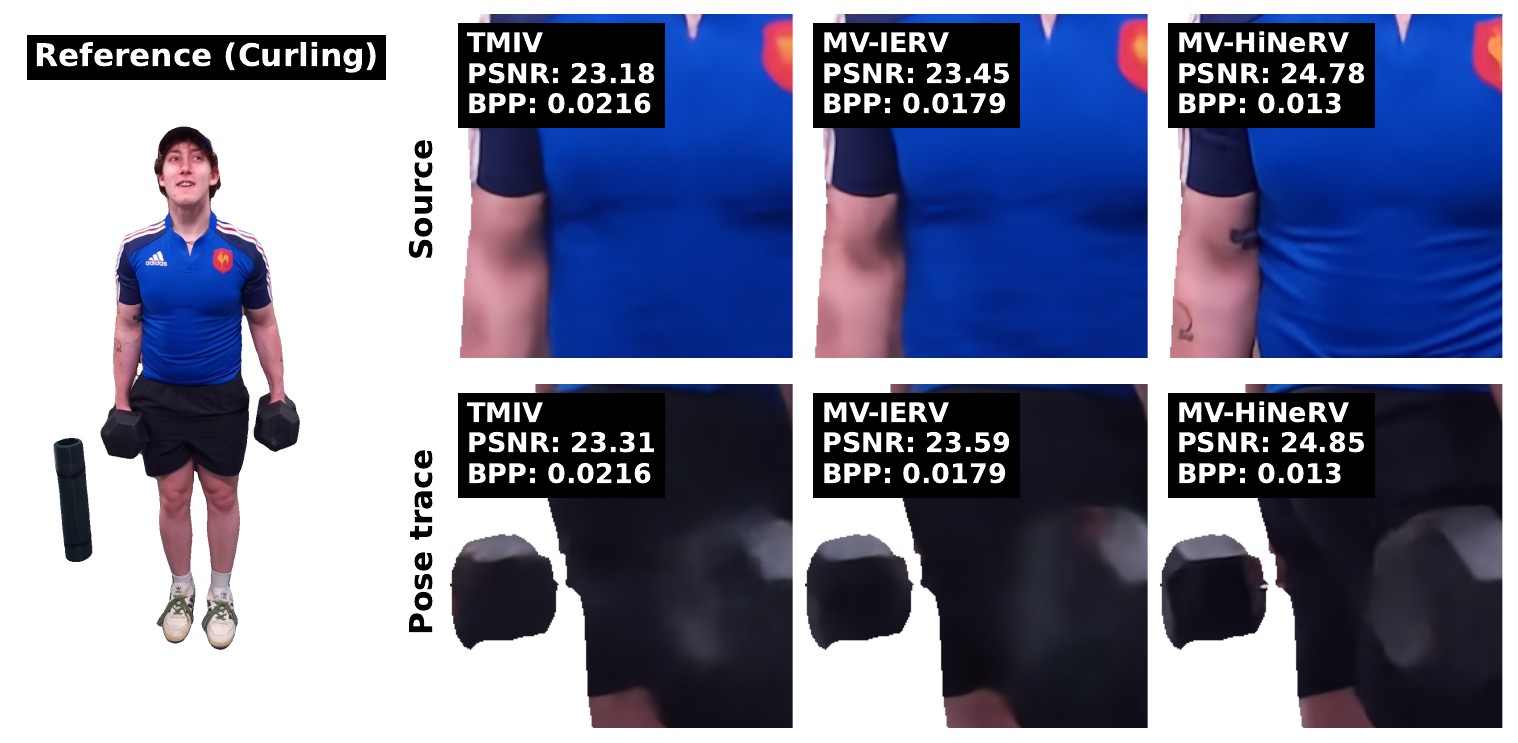}
    \end{minipage}

    \vspace{0.5cm}

    \begin{minipage}{\linewidth}
        \centering
        \includegraphics[width=\linewidth]{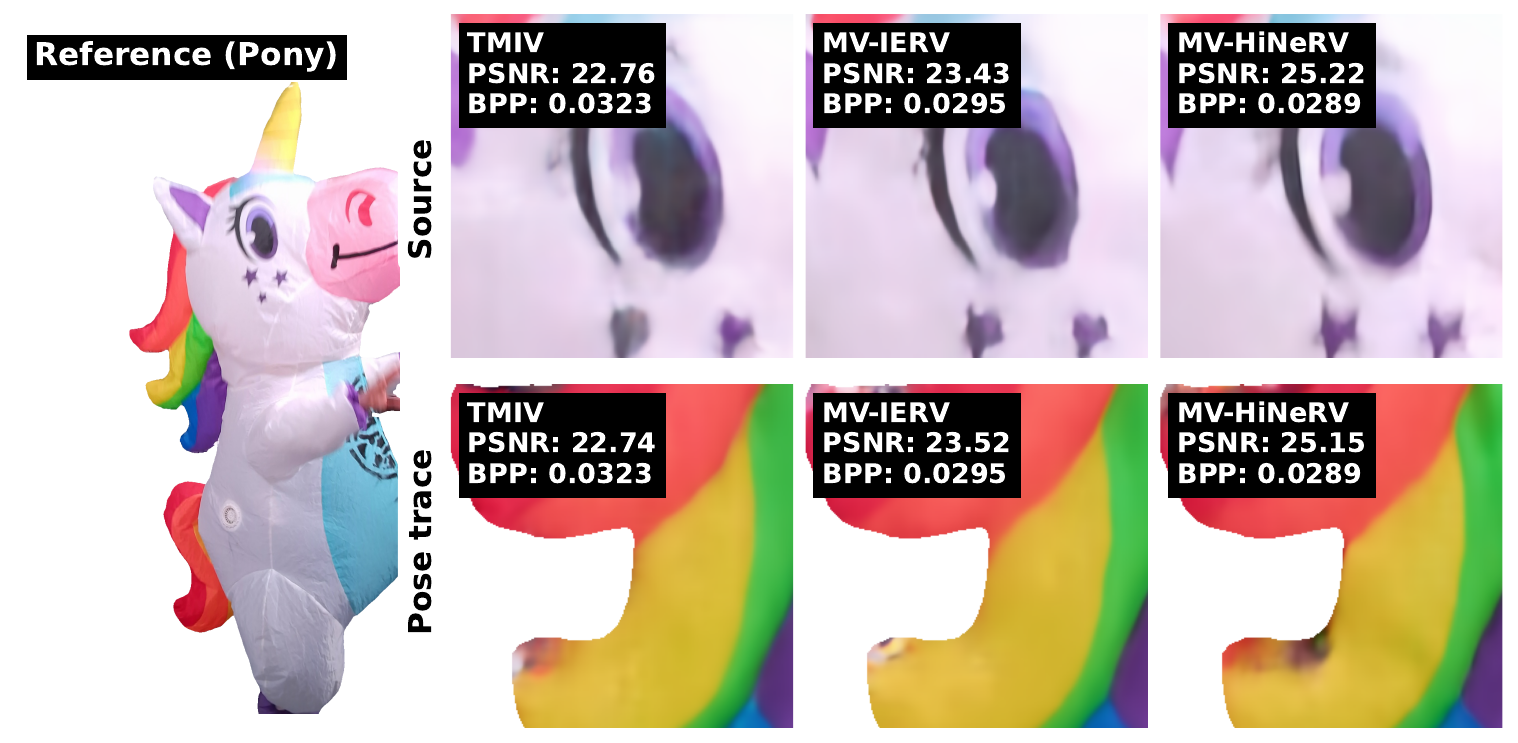}
    \end{minipage}

    \caption{Qualitative comparison of the source and pose trace views by different MVV compression baselines across two scenarios.}
    \label{fig:visu-mvv-compression}
\end{figure}

In \autoref{sec: dnvs.models} we discuss how SC-GS heavily relies on a sparse set of control points to enhance dynamic volumetric surface rigidity; however, from the 512 control points, at most two points are linked to the target performers for most scenes. This is due to the size of the target performer w.r.t the entire scene and the amount of motion. For instance, the Cymbals performer spends a large part of the recording moving around the stage, leading to an almost transparent reconstruction for SC-GS in \autoref{fig: dynamic NVS visual results}. Moreover, the Pony costume is larger (around 8 feet tall) than the human performers, resulting in a larger contribution of control points (5 points) thus a higher degree of motion as shown in \autoref{fig: dynamic NVS temporal results}. Ultimately, this highlights that for sparse-view datasets containing humans, surface-based reconstruction using control points requires further attention. Interestingly, \cite{huang2024sc} offers a separate implementation  of SC-GS that uses masks to improve learning the dynamic control points while also learning static background features. While generating masks for each frame is non-trivial, SC-GS only requires the mask for learning dynamic control points and not for segmentation. Therefore, future work could look at reducing this burden to only a single mask per camera. For a sparse-view dataset like \name, this would be trivial.

Instead, the 4D-GS and STG models are capable of driving motion regardless of view sparsity. However, for STG the visual performance declines after frame 50 as shown in \autoref{fig: dynamic NVS temporal results}. The human performer often fades out of existence, implying that the novel temporal opacity strategy struggles to maintain structure for longer scenes. This also explains why the \cite{li2022neural} dataset in \cite{li2024spacetime} is limited to 50 frames when it is typically tested for 300 frames on other models \cite{li2024spacetime,fridovich2023k,wu20244d}. For \name \ STG produces the best metric results, where the PSNR and SSIM results show a large improvement compared to 4D-GS. 

As all frames contribute equally to the final metric, this shows that PSNR and SSIM are not well suited for evaluating sparse-view dynamic reconstruction quality for long scenes. While the issue is less pronounced with the LPIPS results, it highlights the need for more robust evaluation metrics that can accurately gauge the performance and consistency of 3-D reconstructions w.r.t time. This is challenging to do in 3-D as \name \ contains dynamic and non-rigid textures, so point-based analysis, for example using ball-point query to assess local rigidity \cite{huang2024sc} would not work with water, fire, or smoke. As we discussed, for 2-D this is challenged by temporal bias; therefore, an obvious solution could involve varying the weight of each frame's contribution to the final metric. These weights could be generated based on the speed of motion - for static test cameras, this could be calculated using frame differencing. Otherwise, for scenes involving specific events, the event could be labeled in time and surrounding frames could be weighted w.r.t the time difference. Overall, it is challenging to conceptualize a metric that accounts for both of these things and more, while also being fast, easy to use, and reliable for any dataset. Hence, in future work, we propose to explore a more robust objective metric that is more representative of the perceptual result.

\subsection{MVV Compression}
\autoref{tab:compression-results} shows the BD-rate \cite{BD} results of the benchmarked models, including TMIV \cite{dziembowski2023common}, MV-HiNeRV \cite{kwan2024immersive}, and MV-IERV \cite{zhu2025implicit}. Among these benchmarked codecs, MV-HiNeRV consistently demonstrates superior rate-distortion performance compared to the others across all sequences and bitrates measured and for both source and pose trace views, yielding up to 46.63\% and 48.02\% bit-savings, measured in BD-rate using PSNR and IV-PSNR respectively, over TMIV. This is further supported by its pose trace view synthesis results, as illustrated in \autoref{fig:visu-mvv-compression}. While all baselines are capable of producing novel views with relatively good textural fidelity and geometric consistency, MV-HiNeRV stands out in retaining fine-grained details and fewer visual artifacts. Moreover, as shown in \autoref{tab:compression-results}, the INR-based codecs MV-HiNeRV and MV-IERV exhibit relatively low complexity and achieve decoding speeds higher than those of TMIV, where MV-IERV outperforms MV-HiNeRV in terms of decoding FPS. This suggests that implicit neural representations can offer both high compression efficiency and practical runtime performance, making them strong candidates for real-time or resource-constrained MVV applications. However, it could also be seen from \autoref{fig:visu-mvv-compression} that, at low bitrates, the artifacts are more noticeable for scenes like `Pony' that are associated with high textural complexity. This highlights the value of our dataset in providing diverse and difficult scenarios that effectively stress-test codec performance and expose limitations in current methods.

\section{Conclusion}
\label{sec:conclusion}
This paper introduces \name, a multi-view video dataset with per-frame RGB, depth, and calibration data for dynamic reconstruction and video compression; captured by industry experts in a professional studio. The content is  diverse, including but not limited to age, sex, clothing and actions. This is extended to include various dynamic textures, transparent materials, multi-person scenes and human-animal interactions. The dataset is evaluated on SotA dynamic GS methods and highlights the need for more research on sparse-view dynamic reconstruction. Our findings show that while reconstruction is achievable, the SotA is not capable of reproducing fine details relating to both spatial and temporal features. Moreover, the novel view synthesis experiments reveal discrepancies between the objective metric (PSNR, SSIM, LPIPS-Alex and LPIPS-VGG) and subjective visual results. This occurs due to temporal bias, as a high initial metric result (e.g. during the first 50 frames) may out-weight lower results later on. The paper also benchmarks SotA multi-view video compression methods and validates the preferability of INR-based mutli-view video compression methods in terms of both coding efficiency and coding speed. Finally, to further support current and future research, \name \  is provided in raw form with a set of GUIs to make pre-processing and generating additional data, like point clouds and masks, trivial. We hope this encourages more research on resolving real paradigms, like the sparse, 6DoF, multi-view reconstruction and compression paradigms presented in this paper.




\small
\bibliographystyle{IEEEtran}
\bibliography{mainbib}

\begin{IEEEbiography}[{\includegraphics[width=1in,height=1.25in,clip,keepaspectratio]{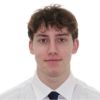}}]{Adrian Azzarelli} received the M.Eng degree in Electronic Engineering with Artificial Intelligence from the University of Southampton, United Kingdom, in 2022. He is currently a Ph.D student in Intelligent Cinematography at the University of Bristol, United Kingdom. His research interests include 3-D video reconstruction and creative applications of computer vision.
\end{IEEEbiography}

\begin{IEEEbiography}[{\includegraphics[width=1in,height=1.25in,clip,keepaspectratio]{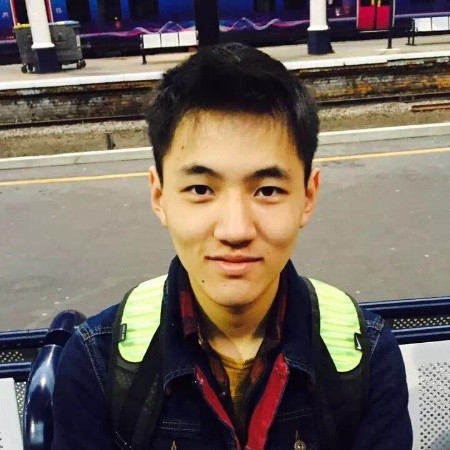}}]{Ge Gao} received the B.Eng degree in Electrical and Electronic Engineering from the University of Manchester in 2018 and M.Sc. degree in Artificial Intelligence from the University of Southampton in 2019. He is currently a Research Associate with the School of Computer Science, University of Bristol. His research interests focus on low-level computer vision including neural video compression, implicit neural representations, and generative models.
\end{IEEEbiography}

\begin{IEEEbiography}[{\includegraphics[width=1in,height=1.25in,clip,keepaspectratio]{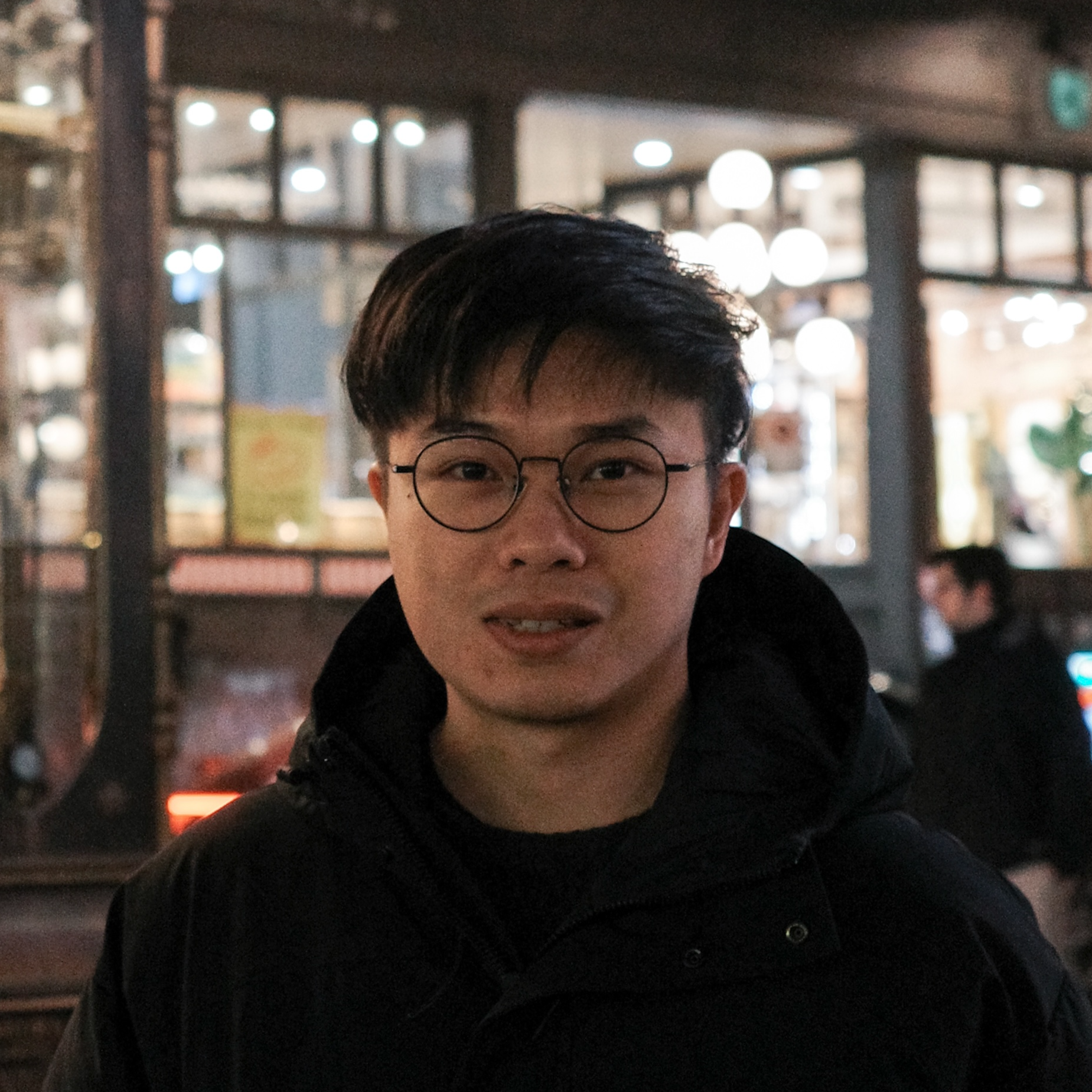}}]{Ho Man Kwan}  received the B.Eng. degree in Computer Engineering and the M.Phil. degree in Electronic and Computer Engineering from the Hong Kong University of Science and Technology in 2018 and 2021, respectively. He is currently pursuing the Ph.D. degree at the University of Bristol. His main research interests include neural compression for 2-D and volumetric videos using implicit neural representations.
\end{IEEEbiography}

\begin{IEEEbiography}[{\includegraphics[width=1in,height=1.25in,clip,keepaspectratio]{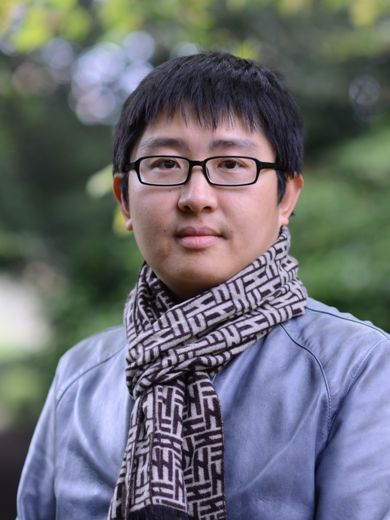}}]{Fan Zhang} (Senior Member, IEEE) received the B.Sc. and M.Sc. degrees from Shanghai Jiao Tong University, Shanghai, China, in 2005 and 2008, respectively, and the Ph.D. degree from the University of Bristol, Bristol, U.K., in 2012. He is currently a Senior Lecturer within the School of Computer Science, University of Bristol. He served as an Associate Editor for IEEE Transactions on Circuits and Systems for Video Technology (2022-2024), and was a guest editor of IEEE Journal on Emerging and Selected Topics in Circuits and Systems (in 2024) and Frontiers in Signal Processing (in 2022). Fan is also a member of the Visual Signal Processing and Communications Technical Committee associated with the IEEE Circuits and Systems Society. His research interests focus on low-level computer vision including video compression, quality assessment, super resolution and video frame interpolation.
\end{IEEEbiography}

\begin{IEEEbiography}[{\includegraphics[width=1in,height=1.25in,clip,keepaspectratio]{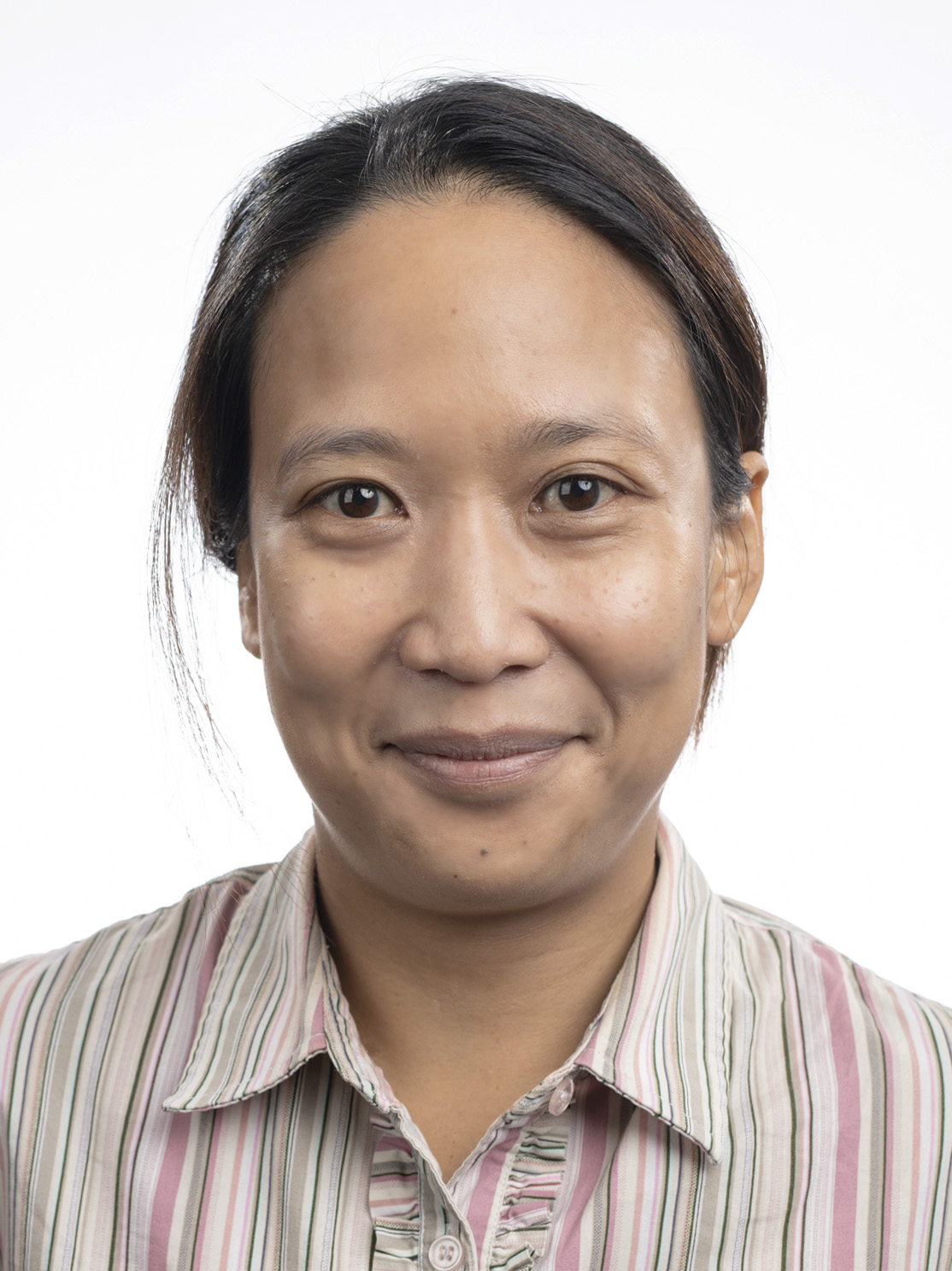}}]{Nantheera Anantrasirichai} (S’04-M’07) received her Ph.D. in Electrical and Electronic Engineering from the University of Bristol. She is currently an Associate Professor in the School of Computer Science at the University of Bristol. Her research interests include image processing and computer vision with challenging data.
\end{IEEEbiography}

\begin{IEEEbiography}[{\includegraphics[width=1in,height=1.25in,clip,keepaspectratio]{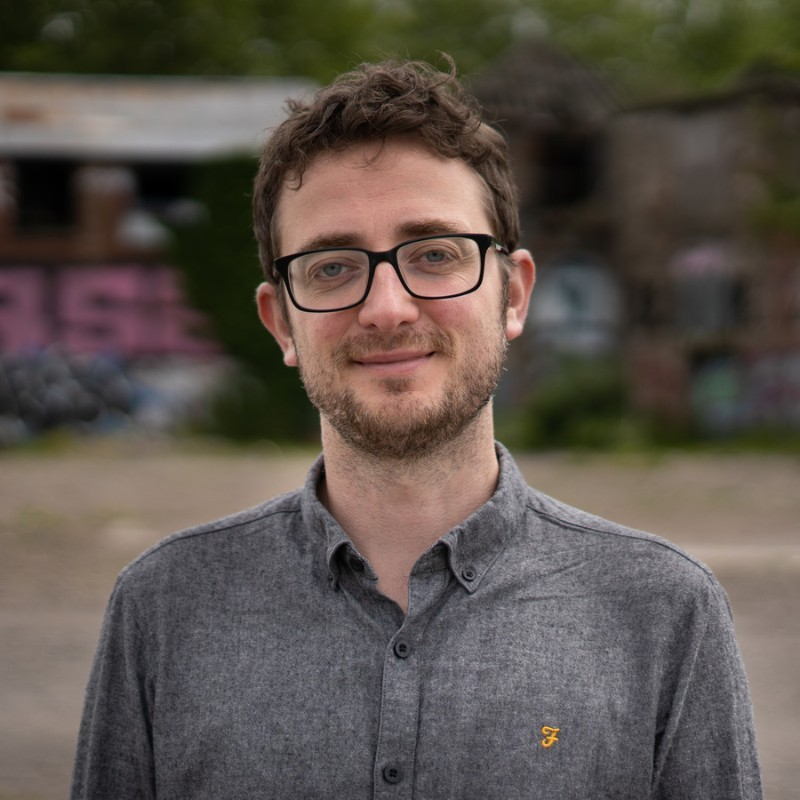}}]{Oliver Moolan-Feroze} received the Ph.D. degree in Computer Vision from the University of Bristol. He is currently the Head of R\&D and a founder at Condense Reality, where he leads research initiatives in immersive technologies. His research interests include 3-D reconstruction, real-time computer vision, and applied deep learning, with a focus on developing practical solutions for emerging spatial computing applications.
\end{IEEEbiography}

\begin{IEEEbiography}[{\includegraphics[width=1in,height=1.25in,clip,keepaspectratio]{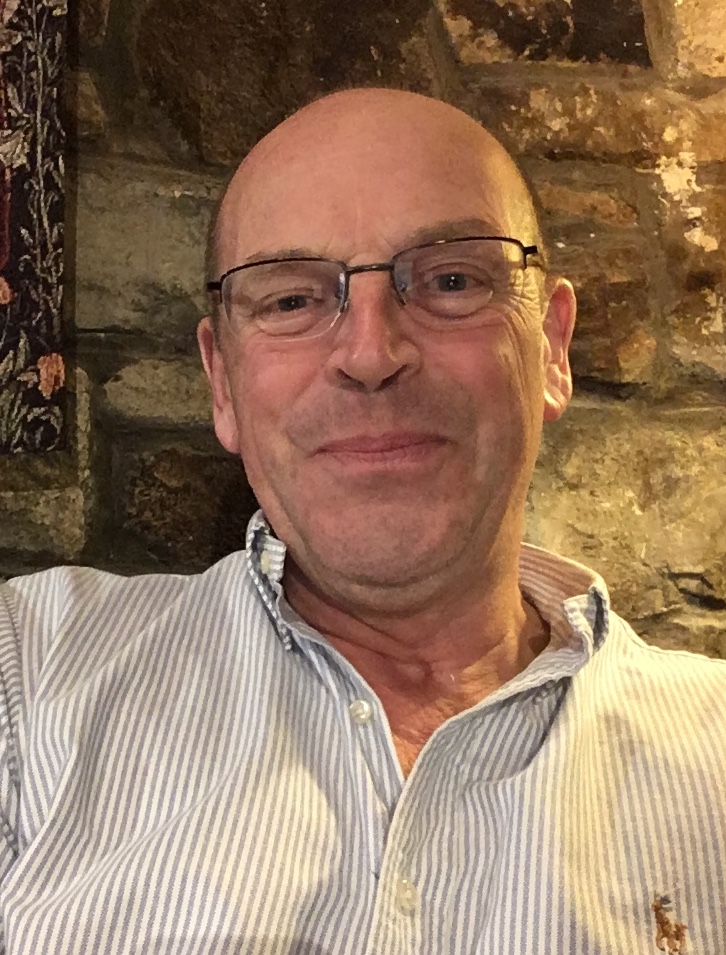}}]{David R. Bull} (Fellow, IEEE) received the B.Sc. degree from the University of Exeter, Exeter, U.K., in 1980, the M.Sc. degree from the University of Manchester, Manchester, U.K., in 1983, and the Ph.D. degree
from the University of Cardiff, Cardiff, U.K., in 1988. He was previously a Systems Engineer with Rolls Royce, Bristol, U.K., and a Lecturer with the University of Wales, Cardiff, U.K. In 1993, he joined the University of Bristol, Bristol, U.K., and is currently its Chair of Signal Processing and the Director of Bristol Vision Institute. He is also the Director of the recently announced £46 m UKRI ‘MyWorld’ Strength in Places Programme. In 2001, he co-founded a university spin-off company, ProVision Communication Technologies Ltd., specializing in wireless video technology. He has authored more than 450 papers on the topics of image and video communications and analysis for wireless, Internet and broadcast applications, together with numerous patents, several of which have been exploited commercially. He is the author of three books, and has delivered numerous invited/keynote lectures and tutorials. He was the recipient of the two IET Premium Awards for his work. Dr. Bull is a Fellow of the Institution of Engineering and Technology.
\end{IEEEbiography}

\end{document}